\documentclass{llncs}
\usepackage{graphicx}
\usepackage{amsmath,amssymb} % define this before the line numbering.
\usepackage{color}
\usepackage{commath}
\usepackage[colorlinks]{hyperref}
\usepackage{mathtools}
\usepackage{nicefrac}
\usepackage{verbatim}
\usepackage{bm}

\usepackage{algorithm}
\usepackage[noend]{algpseudocode}
\makeatletter
\def\BState{\State\hskip-\ALG@thistlm}
\makeatother

\usepackage{textcomp}
\usepackage[detect-all,binary-units]{siunitx}

\usepackage[capitalise]{cleveref}

\newlength\fwidth%
\newlength\fheight%

\usepackage{pgf}
\usepackage{tikz}
\usepackage{pgfplots}
\usetikzlibrary{plotmarks}
\usetikzlibrary{shapes}
\usepgfplotslibrary{units}
\pgfplotsset{compat=newest}
\pgfplotsset{plot coordinates/math parser=false}
\pgfplotsset{try min ticks=3}
\pgfplotsset{max space between ticks=20pt}
\pgfplotsset{invoke before crossref tikzpicture={\tikzexternaldisable},invoke
  after crossref tikzpicture={\tikzexternalenable}}
\pgfplotsset{every x tick label/.append style={font=\tiny, yshift=0.5ex}}
\pgfplotsset{every y tick label/.append style={font=\tiny, xshift=0.5ex}}
\pgfplotsset{every z tick label/.append style={font=\tiny, yshift=0.5ex}}
\tikzset{font={\fontsize{8pt}{12}\selectfont}}
\usetikzlibrary{external}

\DeclareMathOperator*{\argmin}{argmin}

\newcommand{\Argmin}[1]{\underset{#1}{\argmin}}

\newcommand{\T}{^{\mathstrut\scriptscriptstyle{\top}}} % ooh, fancy!
\newcommand{\mc}[1]{\ensuremath{\mathcal{#1}}}   % caligraphic math typeface
\newcommand{\mb}[1]{\ensuremath{\mathbf{#1}}}    % bold math typeface
\newcommand{\mbb}[1]{\ensuremath{\mathbb{#1}}}

\usepackage{siunitx}
\usepackage{microtype}

\usepackage{xspace}
\DeclareRobustCommand\onedot{\futurelet\let\token\onedot}
\def\onedot{\ifx\let\token.\else.\null\fi\xspace}

\def\eg{\emph{e.g}\onedot} 
\def\ie{\emph{i.e}\onedot}

\begin{document}
\pagestyle{headings}
\mainmatter

\title{Photometric Bundle Adjustment for Vision-Based SLAM}

\def\ACCV16SubNumber{554}  % Insert your submission number here

\titlerunning{Photometric BA for VSLAM}
%\authorrunning{ACCV-16 submission ID \ACCV16SubNumber}
\authorrunning{Alismail \and Browning \and Lucey}

\author{Hatem Alismail\thanks{Corresponding author~\href{mailto:halismai@cs.cmu.edu}{halismai@cs.cmu.edu}.} \and Brett Browning \and Simon Lucey}
%\institute{Paper ID \ACCV16SubNumber}
\institute{The Robotics Institute\\Carnegie Mellon University\\%
\texttt{\{halismai,brettb,slucey\}@cs.cmu.edu}}

\maketitle

\begin{abstract}
We propose a novel algorithm for the joint refinement of structure and motion parameters from image data directly without relying on fixed and known correspondences. In contrast to traditional bundle adjustment (BA) where the optimal parameters are determined by minimizing the reprojection error using tracked features, the proposed algorithm relies on maximizing the photometric consistency and estimates the correspondences implicitly. Since the proposed algorithm does not require correspondences, its application is not limited to corner-like structure; any pixel with nonvanishing gradient could be used in the estimation process. Furthermore, we demonstrate the feasibility of refining the motion and structure parameters simultaneously using the photometric in unconstrained scenes and without requiring restrictive assumptions such as planarity. The proposed algorithm is evaluated on range of challenging outdoor datasets, and it is shown to improve upon the accuracy of the state-of-the-art VSLAM methods obtained using the minimization of the reprojection error using traditional BA as well as loop closure.
\end{abstract}

\section{Introduction}
Photometric, or image-based, minimization is a fundamental tool in a myriad of applications such as: optical flow~\cite{Sun2010}, scene flow~\cite{SceneFlow}, and stereo~\cite{Seitz2006,mvs}. Its use in vision-based 6DOF motion estimation has recently been explored demonstrating good results~\cite{lsdslam,kerl13icra,steinbrucker2011real,Meilland13}. Minimizing the photometric error, however, has been limited to frame--frame estimation (visual odometry), or as a tool for depth refinement independent of the parameters of motion~\cite{DTAM}. Consequently, in unstructured scenes, frame--frame minimization of the photometric error cannot reduce the accumulated drift. When loop closure and prior knowledge about the motion and structure are not available, one must resort to the Gold Standard: minimizing the reprojection error using bundle adjustment.

Bundle adjustment (BA) is the problem of jointly refining the parameters of motion and structure to improve a visual reconstruction~\cite{ba}. Although BA is a versatile framework, it has become a synonym to minimizing the reprojection error across multiple views~\cite{Hartley2006,torr00}. The advantages of minimizing the reprojection error are abundant and have been discussed at length in the literature~\cite{Hartley2006,torr00}. In practice, however, there are sources of systematic errors in feature localization that are hard to detect and the value of modeling their uncertainty remains unclear~\cite{kanazawa2001we,brooks2001value}. For example, slight inaccuracies in calibration exaggerate errors~\cite{furukawa2008accurate}, sensor noise and degraded frequency content of the image affect feature localization accuracy~\cite{deriche1990accurate}. Even interpolation artifacts play a non-negligible role~\cite{shimizu2001}. Although minimizing the reprojection is backed by sound theoretical properties~\cite{Hartley2006}, its use in practice must also take into account the challenges and nuances of precisely localizing keypoints~\cite{ba}.

Here, we propose a novel method that further improves upon the accuracy of minimizing the reprojection error and, even state-of-the-art loop closure~\cite{orbslam2015}. The proposed algorithm brings back the image in the loop, and jointly refines the motion and structure parameters to maximize photometric consistency across multiple views. In addition to improved accuracy, the algorithm does not require correspondences. In fact, correspondences are estimated automatically as a byproduct of the proposed formulation.

The ability to perform BA without the need for precise correspondences is attractive because it can enable VSLAM applications where corner extraction is unreliable~\cite{Milford2012}, as well as additional modeling capabilities that extend beyond geometric primitives~\cite{Reid2014,Moreno2013}.

\subsection{Preliminaries and Notation}
\subsubsection{The reprojection error} Given an initial estimate of the scene structure $\set{\bm{\xi}_j}_{j=1}^N$, the viewing parameters per camera $\set{\bm{\theta}_i}_{i=1}^M$, and $\mb{x}_{ij}$ the projection of the $j^\text{th}$ point onto the $i^\text{th}$ the reprojection error is given by
\begin{align}\label{eq:reproj_error}
\epsilon_{ij}(\mb{x}_{ij}; \bm{\theta}_i, \bm{\xi}_j) = \norm{%
\mb{x}_{ij} - \pi\left(\mb{T}(\bm{\theta}_i), \mb{X}(\bm{\xi}_j)\right)},
\end{align}
where $\pi(\cdot,\cdot)$ is the image projection function. The function $\mb{T}(\cdot)$ maps the vectorial representation of motion to a rigid body transformation matrix. Similarly, $\mb{X}(\cdot)$ maps the parameterization of the point to coordinates in the scene.

In this work, we assume known camera calibration parameters as is often the case in VSLAM and parameterize the scene structure using the usual 3D Euclidean coordinates, where $\mb{X}(\bm{\xi}) \coloneqq \bm{\xi}$, and
\begin{align}
\bm{\xi}_j\T = \begin{pmatrix}x_j & y_j & z_j\end{pmatrix}\in\mbb{R}^3.
\end{align}
The pose parameters are represented using twists~\cite{murray1994mathematical}, where the rigid body pose is obtained using the exponential map~\cite{ma2004invitation}, \ie
\begin{align}
\bm{\theta}_i\T \in \mbb{R}^6\quad\text{and}\quad\mb{T}(\bm{\theta}) \coloneqq \exp(\widehat{\bm{\theta}}) \in SE(3).
\end{align}
Our algorithm, similar to minimizing the reprojection error using BA, does not depend on the parameterization. Other representations for motion and structure have been studied in the literature and could be used as well~\cite{hartley2013rotation,civera2008inverse,zhao2015parallaxba}.

\subsubsection{Geometric bundle adjustment} Given an initialization of the scene points and motion parameters, we may obtain a refined estimate by minimizing the squared reprojection error in~\cref{eq:reproj_error} across tracked features, \ie:
\begin{align}\label{eq:ba}
\set{\Delta\bm{\theta}_i^\ast, \Delta\bm{\xi}_j^\ast} = 
\Argmin{\bm{\theta}_i,\bm{\xi}_j}~\sum_{i=1}^M\sum_{j=1}^N \frac{1}{2}\delta_{ij} \epsilon_{ij}^2(\mb{x}_{ij}, \Delta\bm{\theta}_i, \Delta\bm{\xi}_j),
\end{align}
where $\delta_{ij} = 1$ if the $j^\text{th}$ point is visible, or tracked, in the $i^\text{th}$ camera. We call this formulation \emph{geometric} BA.

Minimizing the reprojection error in~\cref{eq:ba} is a large nonlinear optimization problem. Particular to BA is the sparsity pattern of its linearized form, which we can exploit for both large-- and medium--scale problems ~\cite{Hartley2006}.

\subsection{The Photometric Error} The use of photometric information in computer vision has a long and rich history dating back to the seminal works of Lucas and Kanade~\cite{lk} and Horn and Schunk~\cite{horn1981determining}. The problem is usually formulated as a pairwise alignment of two images. One is the reference $\mb{I}_0$, while the other is the input $\mb{I}_1$. The two images are assumed to be related via a parametric transformation. The goal is to estimate the parameters of motion $\bm{p}$ such that the squared intensity error is minimized
\begin{align}
\bm{p}^\ast = \Argmin{\bm{p}}~\sum_{\mb{u} \in \Omega_0}\frac{1}{2}\norm{\mb{I}_0(\mb{u}) - \mb{I}_1(\mb{w}(\mb{u}; \bm{p}))}^2,
\end{align}
where $\mb{u} \in \Omega_0$ denotes a subset of pixel coordinates in the reference image frame, and $\mb{w}\left(\cdot,\cdot\right)$ denotes the warping function~\cite{baker2004lucas}. Minimizing the photometric error has recently resurfaced as a robust solution to visual odometry (VO) from high frame-rate imagery~\cite{engel2015_stereo_lsdslam,kerl13icra,steinbrucker2011real}. Notwithstanding, minimizing the photometric error has not yet been explored for the \emph{joint} optimization of the motion and structure parameters for VSLAM in unstructured scenes.

The proposed approach fills in the gap by providing a photometric formulation for BA, which we call BA \emph{without} correspondences.

\section{Bundle Adjustment Without Correspondences}
\begin{figure}
\centering
\includegraphics[width=0.7\linewidth]{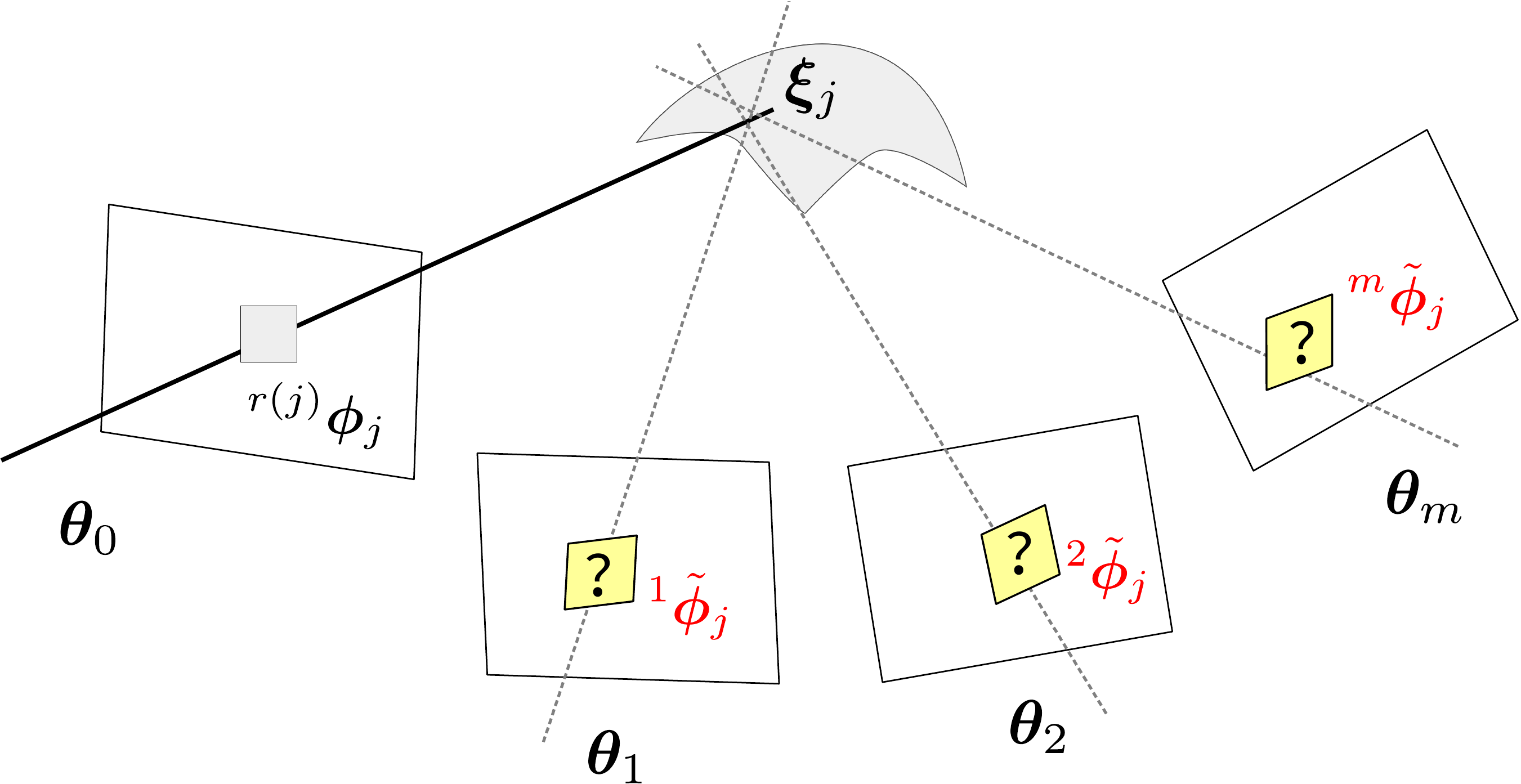}
\caption{{\small Schematic of the proposed approach. We seek to optimize the parameters of motion $\bm{\theta}_i$ and structure $\bm{\xi}_j$ such that the photometric error with respect to a fixed patch at the reference frame is minimized. Correspondences will be estimated implicitly.}}
\label{fig:overview}
\end{figure}

BA is not limited to minimizing the reprojection error~\cite{ba}. We reformulate the problem as follows. First, we assume an initial estimate of the camera poses $\bm{\theta}_i$ as required by geometric BA. However, we do not require tracking information for the 3D points. Instead, for every scene point $\bm{\xi}_j$, we assign a \emph{reference} frame denoted by $r(j)$. The reference frame is used to extract a fixed square patch denoted by $\bm{\phi}_j \in \mbb{R}^D$ over a neighborhood/window denoted by $\mc{N}$. In addition, we compute an initial \emph{visibility} list indicating the frames where the point may be in view. The visibility list for the $j^\text{th}$ point excludes the reference frame and is denoted by:
\begin{align}
\mb{V}_j = \set{k~:~k\ne r(j)\text{ and }\bm{\xi}_j\text{ is visible in frame } k},\text{ for } k \in [1,\ldots,M].
\end{align}
Given this information and the input images $\set{\mb{I}_i}_{i=1}^M$, we seek to estimate an optimal update to the motion ${\Delta\bm{\theta}_i}^\ast$ and structure parameters ${\Delta\bm{\xi}_j}^\ast$ that satisfy
\begin{align}\label{eq:photo_ba}
\set{\Delta\bm{\theta}_i^\ast,\Delta\bm{\xi}_j^\ast} = \Argmin{\Delta\bm{\theta}_i, \Delta\bm{\xi}_j}
\sum_{j=1}^N\sum_{k \in V(j)} \mc{E}(\bm{\phi}_j, \mb{I}_k; \Delta\bm{\theta}_k, \Delta\bm{\xi}_j),
\end{align}
where
\begin{align}\label{eq:photo_error}
\mc{E}(\bm{\phi}, \mb{I}'; \bm{\theta}, \bm{\xi}) =
\sum_{\mb{u} \in \mc{N}}{%
\frac{1}{2}\norm{\bm{\phi}(\mb{u}) - \mb{I}'(\pi(\bm{\theta}, \bm{\xi}) + \mb{u})}^2}.
\end{align}
The notation $\mb{I}'(\pi(\cdot,\cdot) + \mb{u})$ indicates sampling the image intensities in a neighborhood about the current projection of the point. Since image projection results in subpixel coordinates, the image is sampled using an appropriate interpolation scheme (bilinear in this work). The objective for a single point is illustrated schematically in~\cref{fig:overview}.

\subsubsection{Linearization and sparsity}
The optimization problem in~\cref{eq:photo_ba} is nonlinear and its solution proceeds with standard techniques. Let $\mb{\bm{\theta}}$ and $\bm{\xi}$ denote the current estimate of the camera and the scene point, and let the current projected pixel coordinate in the image plane be given by
\begin{align}
\mb{u}'= \pi(\mb{T}(\bm{\theta}),\mb{X}(\bm{\xi})),
\end{align}
then taking the partial derivatives of the $1^\text{st}$-order expansion of the photometric error in~\cref{eq:photo_error} with respect to the motion and structure parameters we obtain:
\begin{align}\label{eq:J_theta}
\frac{\partial\mc{E}}{\partial\bm{\theta}} &=
\sum_{\mb{u}\in\mc{N}}%
\mb{J}\T(\bm{\theta})\left|
\bm{\phi}(\mb{u}) - \mb{I}'(\mb{u}' + \mb{u}) - \mb{J}(\bm{\theta})\Delta\bm{\theta}\right| \\
\label{eq:J_xi}
\frac{\partial\mc{E}}{\partial\bm{\xi}} &=
\sum_{\mb{u}\in\mc{N}}\mb{J}\T(\bm{\xi})\left|
\bm{\phi}(\mb{u}) - \mb{I}'(\mb{u}' + \mb{u}) - \mb{J}(\bm{\xi})\Delta\bm{\xi}\right|,
\end{align}
where $\mb{J}(\bm{\theta}) = \nabla\mb{I}(\mb{u}' + \mb{u})\frac{\partial\mb{u}'}{\partial\bm{\theta}}$, and $\mb{J}(\bm{\xi}) = \nabla\mb{I}(\mb{u}' + \mb{u})\frac{\partial\mb{u}'}{\partial\bm{\xi}}$. The partial derivatives of the projected pixel location with respect to the parameters are identical to those obtained when minimizing the reprojection error in \cref{eq:reproj_error}, and $\nabla\mb{I} \in \mbb{R}^{1\times 2}$ denotes the image gradient. By equating the partial derivatives in~\cref{eq:J_theta,eq:J_xi} to zero we arrive at the normal equations which can be solved efficiently using standard methods~\cite{Nocedal2006NO}.

\begin{figure}
\centering
\includegraphics[width=0.7\linewidth]{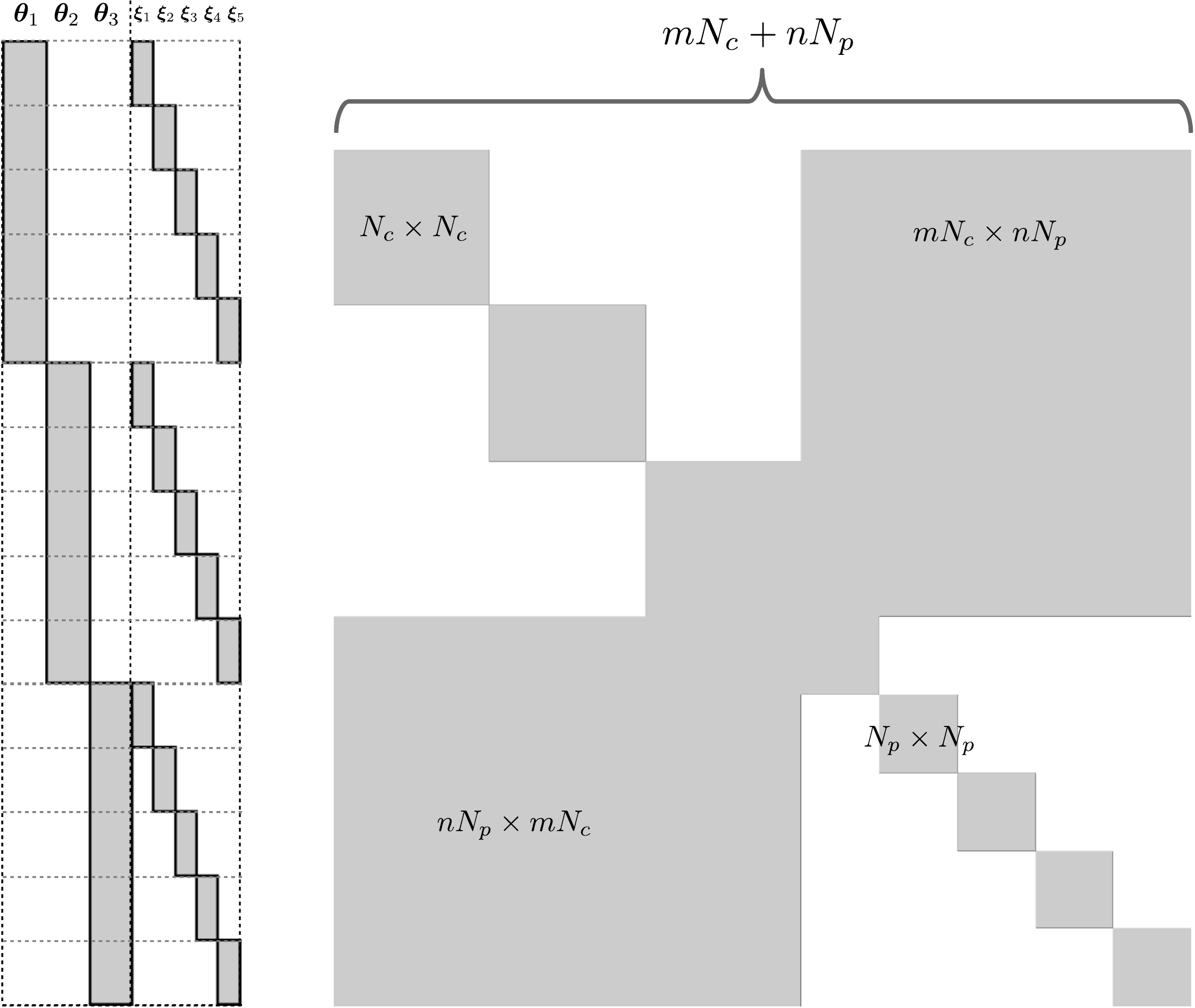}
\caption{{\small Shown on the left is the form of the Jacobian for a photometric bundle adjustment problem consisting of $3$ cameras, $4$ points, and using a $9$-dimensional descriptor, with $N_c=6$ parameters per camera, and $N_p=3$ parameters per point. The form of the normal equations is shown on the right. The illustration is not up to scale across the two figures.}}
\label{fig:sparsity}
\end{figure}

We note that the Jacobian involved in solving the photometric error has a higher dimensionality than its counterpart in geometric BA. This is because the dimensionality of intensity patches ($D \ge 3\times 3$) is usually higher than the dimensionality of feature projections (typically $2$ for a monocular reconstruction problem). Nonetheless, the Hessian remains \emph{identical} to minimizing the reprojection error and the linear system remains sparse and is efficient to decompose. The sparsity pattern of the photometric BA problem is illustrated in~\cref{fig:sparsity}.

Another important note is that since the parameters of motion and structure are refined jointly, the location of the patch at the reference frame $\bm{\phi}(\mb{u})$ in \cref{eq:photo_error} will additionally depend on the pose parameters of the reference frame. Allowing the reference patch to ``move'' during the optimization adds additional terms to the Hessian (additional terms will appear in the motion parameters blocks of the Hessian, these are shown at left hand corner of the Hessian in \cref{fig:sparsity}). In terms of computational complexity, the additional runtime from allowing the reference patch to move is minimal as the algorithm is implemented in a sliding window fashion. However, including inter--pose dependencies is undesirable as, depending on the initialization quality, the location of the reference patch might drift. For instance, we might introduce a biased solution where the patches drift to image regions with brighter absolute intensity values in an attempt to obtain the minimum energy in low-texture areas.

To address this problem, we fix the patch appearance at the reference frame by storing the patch values as soon as the reference frame is selected. This is equivalent to assuming a known patch appearance from an independent source. Under this assumption, the optimization problem now becomes: given a known and fixed patch appearance of a $3$D point in the world, refine the parameters of the structure and motion such that photometric error between the fixed patch and its projection onto the other frames is minimized. This assumption has two advantages: (1) the Hessian sparsity pattern remains identical to the familiar form when minimizing the reprojection error using traditional BA, and (2) we can refine the three coordinates (or the full four projective coordinates~\cite{ba}) of the scene points as opposed to only refining depth along a fixed ray in space.

In addition to improving the accuracy of VSLAM, the algorithm does not require extensive parameter tuning. This is now possible by allowing the algorithm to determine the correct correspondences, hence eliminating the many steps required to ensure outlier-free correspondences with traditional BA. The current implementation of the proposed algorithms is controlled by the three parameters summarized in~\cref{table:params} and explained next.

\begin{table}
\centering
\caption{{\small Configuration parameters for the proposed algorithm shown in \cref{myalg}.}}
\label{table:params}
\begin{tabular}{p{1.8in}c}
\hline\noalign{\smallskip}
Parameter & Value\\
\noalign{\smallskip}
\hline
\noalign{\smallskip}
Patch radius & $1$ or $2$ \\
Non maxima suppression radius & $1$ \\
Max distance to update $V_j$ & $2$ \\
\hline
\end{tabular}
\end{table}

\subsubsection{Selecting pixels} While it is possible to select pixel locations at every frame using a standard feature detector, such as Harris~\cite{harris} or FAST~\cite{rosten2006machine}, we opt to use a simpler and more efficient strategy based on the gradient magnitude of the image.  This is performed by selecting pixels with a local maxima in a $3\times 3$ neighborhood of the absolute gradient magnitude of the image. The rationale is that pixels with vanishing intensity gradients do not contribute to the linear system in~\cref{eq:J_theta,eq:J_xi}. Hence, pixels with larger gradients are preferable since they indicate a measure of textureness~\cite{klt}. Other strategies for pixel selection could used~\cite{dellaert2000structure,meilland2010}, but we found that the current scheme works well as it ensures an even distribution of coordinates across the field-of-view of the camera~\cite{nister04}. The proposed pixel selection strategy is also beneficial as it is not restricted to corner-like structure and allows us to use pixels from low-texture areas. We note that this pixel selection step selects pixels at integer locations; there is no need to compute accurate subpixel positions of the selected points at this stage.

In image-based (photometric) optimization there is always a distinguished reference frame providing fixed measurements~\cite{Irani2000,SteinShashua2000,DTAM}. Selecting a single reference in photometric VSLAM is unnecessary and may be inadvisable. It is unnecessary as the density of reconstruction is not our main goal. It is inadvisable because we need the scene points to serve as tie points~\cite{agouris1996automated} and form a strong network of constraints~\cite{ba}. Given the nature of camera motion in VSLAM selecting points from every frame ensures the strong network of connections between the tie points. For instance, typical hand-held and ground robots motions are mostly forward with points leaving the field-of-view rapidly.

Selecting new scene points at every frame using the aforementioned non maxima suppression procedure has one caveat. If we always select pixels with strong gradients between consecutive frames, then we are likely to track previous scene points rather than finding new ones. This is because pixels with locally maximum gradient magnitude at the consecutive frame are most likely images of previously selected points. Treating projections of previously initialized scene points as new observations is problematic because it introduces unwanted dependencies in the normal equations and superficially increases the number of independent measurements in the linearized system of equations.

\begin{figure}
\centering
\includegraphics[width=0.35\linewidth]{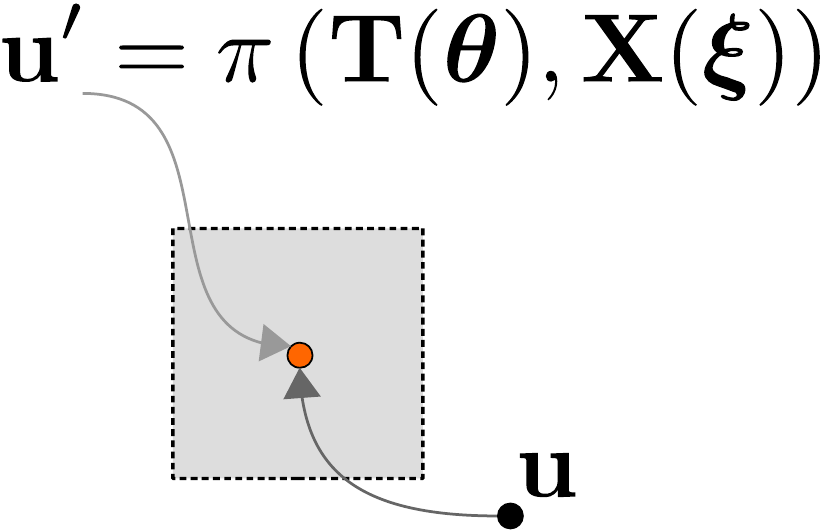}
\caption{{\small Illustration of how we avoid reinitializing the same point at a new frame. Using the pose initialization of the new frame $\bm{\theta}$, we project previous scene points and reserve a $3\times 3$ area where new scene points cannot be initialized.}}
\label{fig:blockmask}
\end{figure}

To address this issue, we assume that the scene and motion initialization is accurate enough to predict the location of current scenes in the new frame. Prior to initializing new scene points, we use the provided pose initialization to warp all previously detected scene points that are active in the optimization sliding window onto the new frame. After that, we mark a $3\times 3$ square area at the projection location of the previous scene points as an invalid location for selecting new points. This step is illustrated in~\cref{fig:blockmask}, and is best summarized in our pseudo code shown in~\cref{myalg}.

The number of selected points per frame varies depending on the image resolution and texture information in the image. In our experiments, this number ranges between $\approx 4000\text{--}10000$ points per image.

\subsubsection{Determining visibility} Ideally, we would like to assume that newly initialized scene points are visible in all frames and to rely on the algorithm to reliably determine if this is the case. However, automatically determining the visibility information along with structure and motion parameters is challenging, as many scene points quickly go out of view, or become occluded. Their inclusion in the optimization problem incurs an unnecessary computational complexity, reduces robustness, and increases the uncertainty of the estimated parameters.

An efficient and reliable measure to detect occlusions and points that cannot be matched reliably is the normalized correlation. For all scene points that are close to the current frame $i$, we use the pose initialization $\mb{T}_i$ to extract a $5\times 5$ intensity patch. The patch is obtained by projecting the scene points to the new frame and its visibility list is updated if the zero-mean normalized correlation score (ZNCC) is greater than $0.6$. We allow $\pm{2}$ frames for a point to be considered close, \ie $|i - r(j)|\le 2$. This procedure is similar to determining the visibility information in multi-view stereo algorithms~\cite{mvs} and is best summarized in~\cref{myalg}.

\begin{algorithm}
\caption{Summary of image processing in our algorithm}
\label{myalg}
\begin{algorithmic}[1]
\Procedure{ProcessFrame}{$\mb{I}_i, \mb{T}_i$}
  \State\text{\textbf{Step 1}: establish connections to the new frame}
  \State \textbf{mask} = \texttt{all\_valid}(\texttt{rows}($\mb{I}$), \texttt{cols}($\mb{I}$))
  \For{all scene points $\mb{X}_j$ in sliding window}
    \If{\text{reference frame $r(j)$ is too far from $i$}}
    \State\texttt{continue}
    \EndIf
    \State \text{$\mb{x} \coloneqq$ projection of $\mb{X}_j$ onto image $\mb{I}_i$ using pose $\mb{T}_i$}
    \State \text{$\bm{\phi}' \coloneqq$ patch at $\mb{x}$ and $\bm{\phi}\coloneqq$ reference patch for $\mb{X}_j$}
    \If {zncc($\bm{\phi}$, $\bm{\phi}') >$\text{ threshold}}
        \State \text{add frame $i$ to visibility list $V_j$}
        \State \text{\textbf{mask}($\mb{u}$) = \texttt{invalid}}
    \EndIf
  \EndFor
  \Statex
  \State\text{\textbf{Step 2}: add new scene points}
  \State\text{$\mb{G}\coloneqq$ gradient magnitude of $\mb{I}_i$}
  \For{all pixels $\mb{u}$ in $\mb{I}_i$}
  \If{$\mb{u}$ is a local maxima in $\mb{G}$}
    \If{location $\mb{u}$ is \texttt{valid} in $\textbf{mask}$}
      \State\text{initialize a new point $\mb{X}$ with reference patch at $\mb{I}(\mb{u})$}
    \EndIf
  \EndIf
  \EndFor
\EndProcedure
\end{algorithmic}
\end{algorithm}

\subsubsection{Optimization details}
We use the Ceres optimization library~\cite{ceres-solver}, designed for BA problems, to optimize the objective in~\cref{eq:photo_ba}. We use the Levenberg-Marquardt algorithm~\cite{levenberg44,marquardt63} to minimize a Huber loss function instead of squared loss to improve robustness. Termination tolerances are set to $1\times 10^{-6}$, and automatic differentiation facilities are used. The image gradients used in the linearized system in~\cref{eq:J_theta,eq:J_xi} are computed using a central-difference filter given by $\frac{1}{2}\left[-1,~0,1\right]$. Finally, we make explicit use of the Schur complement to obtain a more efficient solution.

Since scene points do not remain in view for an extended period in most VSLAM datasets, the photometric refinement step is performed using a sliding window of five frames~\cite{Sunderhauf2006}. The motion parameters of the first frame in the sliding window is held constant to fixate the Gauge freedom~\cite{ba}. The 3D parameters of the scene points in the first frame, however, are included in the optimization.

\begin{comment}
\begin{table}
\centering
\caption{\small Nonlinear optimization parameters using Ceres~\cite{ceres-manual} v.~1.12.0.}
\label{table:ceres}
\begin{tabular}{p{1.5in}l}
\hline\noalign{\smallskip}
Parameter & Value\\
\noalign{\smallskip}
\hline
\noalign{\smallskip}
Algorithm     & Levenberg-Marquardt \\
Precondition  & Cluster Jacobi (block diagonal)~\cite{kushal2012visibility}\\
Visibility clustering & Single linkage\\
Linear solver & Sparse Schur complement~\cite{ba}\\
\hline
\end{tabular}
\end{table}
\end{comment}

\section{Experiments}
In this section, we evaluate the performance of the proposed algorithm on two commonly used VSLAM benchmarks to facilitate comparisons with the state-of-the-art. The first is the KITTI benchmark~\cite{Geiger2012CVPR}, which contains imagery from an outdoor stereo camera mounted on a vehicle. The second is the Malaga dataset~\cite{blanco2013mlgdataset}, which is particularly challenging for VSLAM because the baseline of the camera ($12$cm) is small relative to the scene structure.

\subsection{The KITTI Benchmark}
\subsubsection{Initializing with geometric BA}
Torr and Zisserman~\cite{torr00} convincingly argue that the estimation of structure and motion should proceed by feature extraction and matching to provide a good initialization for BA-based refinement techniques. Here, we use the output of ORB-SLAM~\cite{orbslam2015}, a recently proposed state-of-the-art VSLAM algorithm, to initialize our method. ORB-SLAM not only performs geometric BA, but also implements loop closure to reduce drift. The algorithm is currently one of the top performing algorithms on the KITTI benchmark~\cite{Geiger2012CVPR}.

We only use the pose initialization from ORB-SLAM. We do not make use of the refined 3D points as they are available at selected keyframes only. This is because images in the KITTI benchmark are collected at \SI{10}{\Hz}, while the vehicle speed exceeds \SI{80}{\km/\hour} in some sections. Subsequently, the views are separated by a large baseline, which violates the small displacement assumption required for the validity of linearization in~\cref{eq:J_theta,eq:J_xi}.

Hence, to initialize 3D points we use the standard block matching stereo algorithm implemented in OpenCV. This is a winner-takes-all brute force search strategy based on the sum of absolute intensity differences (SAD). The algorithm is configured to search for $128$ disparities using a $7\times 7$ aggregation window and a left--right consistency check.

The choice of initializing the algorithm with ORB-SLAM is intentional to assess the accuracy of the algorithm in comparison to the Gold Standard solution from traditional BA. We note, however, that we could use LSD-SLAM~\cite{lsdslam} to obtain a VSLAM system without correspondences at all. In fact, initial pose estimates could be provided by external sensors, such as low quality GPS.

\begin{figure}
\centering
\setlength\fwidth{0.8\linewidth}
\setlength\fheight{3.0in}
\definecolor{mycolor1}{rgb}{0.62745,0.12549,0.94118}%
\definecolor{mycolor2}{rgb}{1.00000,0.00000,1.00000}%
\begin{tikzpicture}

\begin{axis}[%
width=0.411\fwidth,
height=0.419\fheight,
at={(0\fwidth,0.581\fheight)},
scale only axis,
xmin=100,
xmax=800,
xlabel={{\tiny Path Length [m]}},
xmajorgrids,
ymin=0.8319,
ymax=1.5367,
ylabel={{\tiny Translation [\%]}},
ymajorgrids,
axis background/.style={fill=white}
]
\addplot [color=white!30!black,dashdotted,line width=0.75pt,mark=o,mark options={solid},forget plot]
  table[row sep=crcr]{%
100	1.5367\\
200	1.3884\\
300	1.3259\\
400	1.2838\\
500	1.2492\\
600	1.2135\\
700	1.1393\\
800	1.1043\\
};
\addplot [color=blue,dashed,line width=0.75pt,mark=triangle,mark options={solid},forget plot]
  table[row sep=crcr]{%
100	1.5061\\
200	1.2913\\
300	1.2023\\
400	1.1309\\
500	1.0697\\
600	1.01\\
700	0.886\\
800	0.8319\\
};
\addplot [color=black,solid,line width=0.75pt,mark=triangle,mark options={solid,rotate=180},forget plot]
  table[row sep=crcr]{%
100	1.364\\
200	1.1944\\
300	1.1444\\
400	1.1302\\
500	1.1502\\
600	1.1786\\
700	1.1973\\
800	1.2424\\
};
\addplot [color=mycolor1,dotted,line width=0.75pt,mark=+,mark options={solid},forget plot]
  table[row sep=crcr]{%
100	1.4119\\
200	1.2858\\
300	1.234\\
400	1.2033\\
500	1.1765\\
600	1.1493\\
700	1.1005\\
800	1.0985\\
};
\addplot [color=mycolor2,solid,line width=0.75pt,mark=square,mark options={solid},forget plot]
  table[row sep=crcr]{%
100	1.41139761725242\\
200	1.22256968871482\\
300	1.13928281492188\\
400	1.07262627677637\\
500	1.0171193133773\\
600	0.962885625755284\\
700	0.919859367644675\\
800	0.867981508902624\\
};
\end{axis}

\begin{axis}[%
width=0.411\fwidth,
height=0.419\fheight,
at={(0.54\fwidth,0.581\fheight)},
scale only axis,
xmin=100,
xmax=800,
xlabel={Path Length [m]},
xmajorgrids,
ymin=0.00129985994442058,
ymax=0.00544309905374282,
ylabel={{\tiny Rotation Error [deg/m]}},
ymajorgrids,
axis background/.style={fill=white},
legend style={legend cell align=left,align=left,draw=white!15!black}
]
\addplot [color=white!30!black,dashdotted,line width=0.75pt,mark=o,mark options={solid}]
  table[row sep=crcr]{%
100	0.004870141258612\\
200	0.00360963410932419\\
300	0.00309397209370645\\
400	0.00269290163711487\\
500	0.00240642273954946\\
600	0.00223453540101021\\
700	0.00200535228295788\\
800	0.00183346494441863\\
};
\addlegendentry{{\tiny MFI}};

\addplot [color=blue,dashed,line width=0.75pt,mark=triangle,mark options={solid}]
  table[row sep=crcr]{%
100	0.004870141258612\\
200	0.00349504255029802\\
300	0.00280749319614103\\
400	0.00234912696003638\\
500	0.00200535228295788\\
600	0.00171887338539247\\
700	0.00143239448782706\\
800	0.00131780292880089\\
};
\addlegendentry{{\tiny ORB-SLAM}};

\addplot [color=black,solid,line width=0.75pt,mark=triangle,mark options={solid,rotate=180}]
  table[row sep=crcr]{%
100	0.00544309905374282\\
200	0.0038961130068896\\
300	0.00332315521175877\\
400	0.0029220847551672\\
500	0.00263560585760179\\
600	0.00246371851906254\\
700	0.00223453540101021\\
800	0.00211994384198405\\
};
\addlegendentry{{\tiny LSD-SLAM}};

\addplot [color=mycolor1,dotted,line width=0.75pt,mark=+,mark options={solid}]
  table[row sep=crcr]{%
100	0.00481284547909891\\
200	0.0035523383298111\\
300	0.00303667631419336\\
400	0.00263560585760179\\
500	0.00234912696003638\\
600	0.00217723962149713\\
700	0.00200535228295788\\
800	0.00183346494441863\\
};
\addlegendentry{{\tiny VoBA}};

\addplot [color=mycolor2,solid,line width=0.75pt,mark=square,mark options={solid}]
  table[row sep=crcr]{%
100	0.00460717732339519\\
200	0.00318769869346741\\
300	0.00250412493404579\\
400	0.00209072433245846\\
500	0.00179016321478705\\
600	0.00158964760155725\\
700	0.00141761991516839\\
800	0.00129985994442058\\
};
\addlegendentry{{\tiny Ours}};

\end{axis}

\begin{axis}[%
width=0.411\fwidth,
height=0.419\fheight,
at={(0\fwidth,0\fheight)},
scale only axis,
xmin=14.4,
xmax=86.4,
xlabel={Speed [km/h]},
xmajorgrids,
ymin=0.898075207381531,
ymax=3.0,
ylabel={{\tiny Translation Error [\%]}},
ymajorgrids,
axis background/.style={fill=white}
]
\addplot [color=white!30!black,dashdotted,line width=0.75pt,mark=o,mark options={solid},forget plot]
  table[row sep=crcr]{%
14.4	1.4829\\
21.6	1.3059\\
28.8	1.1062\\
36	  1.0599\\
43.2	1.7691\\
50.4	2.1804\\
57.6	1.4184\\
64.8	1.1532\\
72	  1.7349\\
79.2	1.8058\\
86.4	2.0601\\
};
\addplot [color=blue,dashed,line width=0.75pt,mark=triangle,mark options={solid},forget plot]
  table[row sep=crcr]{%
14.4	1.8962\\
21.6	1.1688\\
28.8	0.9283\\
36	  0.9182\\
43.2	1.9654\\
50.4	2.3368\\
57.6	1.3393\\
64.8	1.4132\\
72	  1.5209\\
79.2	1.5316\\
86.4	1.6692\\
};
\addplot [color=black,solid,line width=0.75pt,mark=triangle,mark options={solid,rotate=180},forget plot]
  table[row sep=crcr]{%
14.4	1.2656\\
21.6	1.0802\\
28.8	0.9499\\
36	  0.9216\\
43.2	1.3668\\
50.4	1.6497\\
57.6	1.5664\\
64.8	1.595\\
72	  1.8725\\
79.2	1.9601\\
86.4	2.8888\\
};
\addplot [color=mycolor1,dotted,line width=0.75pt,mark=+,mark options={solid},forget plot]
  table[row sep=crcr]{%
14.4	1.4866\\
21.6	1.3061\\
28.8	1.0721\\
36	  1.0154\\
43.2	1.8327\\
50.4	2.3421\\
57.6	1.5775\\
64.8	1.123\\
72	  1.4876\\
79.2	1.5178\\
86.4	1.3887\\
};
\addplot [color=mycolor2,solid,line width=0.75pt,mark=square,mark options={solid},forget plot]
  table[row sep=crcr]{%
14.4	0.898075207381531\\
21.6	1.25099644828612\\
28.8	0.905938946220056\\
36	  1.07437315160973\\
43.2	1.19735185971238\\
50.4	1.26030509015547\\
57.6	2.3973683940959\\
64.8	2.23981065491843\\
72	  1.71466012564139\\
79.2	2.07877656549367\\
86.4	2.64264318205157\\
};
\end{axis}

\begin{axis}[%
width=0.411\fwidth,
height=0.419\fheight,
at={(0.54\fwidth,0\fheight)},
scale only axis,
xmin=14.4,
xmax=86.4,
xlabel={Speed [km/h]},
xmajorgrids,
ymin=0.000916732472209317,
ymax=0.00676090198254371,
ylabel={{\tiny Rotation Error [deg/m]}},
ymajorgrids,
axis background/.style={fill=white}
]
\addplot [color=white!30!black,dashdotted,line width=0.75pt,mark=o,mark options={solid},forget plot]
  table[row sep=crcr]{%
14.4	0.00624523996692597\\
21.6	0.00366692988883727\\
28.8	0.00303667631419336\\
36	  0.00269290163711487\\
43.2	0.00303667631419336\\
50.4	0.00378152144786343\\
57.6	0.0029220847551672\\
64.8	0.00143239448782706\\
72	  0.0025783100780887\\
79.2	0.00269290163711487\\
86.4	0.00217723962149713\\
};
\addplot [color=blue,dashed,line width=0.75pt,mark=triangle,mark options={solid},forget plot]
  table[row sep=crcr]{%
14.4	0.00630253574643906\\
21.6	0.00303667631419336\\
28.8	0.00246371851906254\\
36	  0.00234912696003638\\
43.2	0.00360963410932419\\
50.4	0.00481284547909891\\
57.6	0.0035523383298111\\
64.8	0.00269290163711487\\
72	  0.00269290163711487\\
79.2	0.00263560585760179\\
86.4	0.00280749319614103\\
};
\addplot [color=black,solid,line width=0.75pt,mark=triangle,mark options={solid,rotate=180},forget plot]
  table[row sep=crcr]{%
14.4	0.00676090198254371\\
21.6	0.00366692988883727\\
28.8	0.00297938053468028\\
36	  0.00275019741662795\\
43.2	0.00383881722737652\\
50.4	0.0055003948332559\\
57.6	0.00532850749471666\\
64.8	0.00338045099127186\\
72	  0.00366692988883727\\
79.2	0.00366692988883727\\
86.4	0.00360963410932419\\
};
\addplot [color=mycolor1,dotted,line width=0.75pt,mark=+,mark options={solid},forget plot]
  table[row sep=crcr]{%
14.4	0.00658901464400447\\
21.6	0.00372422566835035\\
28.8	0.00297938053468028\\
36	  0.00263560585760179\\
43.2	0.00326585943224569\\
50.4	0.00395340878640268\\
57.6	0.00320856365273261\\
64.8	0.00137509870831398\\
72	  0.00246371851906254\\
79.2	0.00252101429857562\\
86.4	0.000916732472209317\\
};
\addplot [color=mycolor2,solid,line width=0.75pt,mark=square,mark options={solid},forget plot]
  table[row sep=crcr]{%
14.4	0.00472058038312473\\
21.6	0.00380309756873639\\
28.8	0.00243985186505357\\
36	  0.00231635644717639\\
43.2	0.00197079002227384\\
50.4	0.00216974119395124\\
57.6	0.00383545600268296\\
64.8	0.00306214754017764\\
72	  0.00227661095794878\\
79.2	0.00182502391280481\\
86.4	0.00139421417910356\\
};
\end{axis}
\end{tikzpicture}%
\caption{{\small Comparison to state-of-the-art algorithms on the KITTI benchmark. Our approach performs the best. Error in our approach correspond to segments of the data when the vehicle is driving at a high speed, which increases the magnitude of motion between frames and affects the linearization assumptions. No loop closure, or keyframing is performed using our algorithm. Improvement is shown qualitatively in \cref{fig:kitti_00_path}.}}
\label{fig:ba_results}
\end{figure}
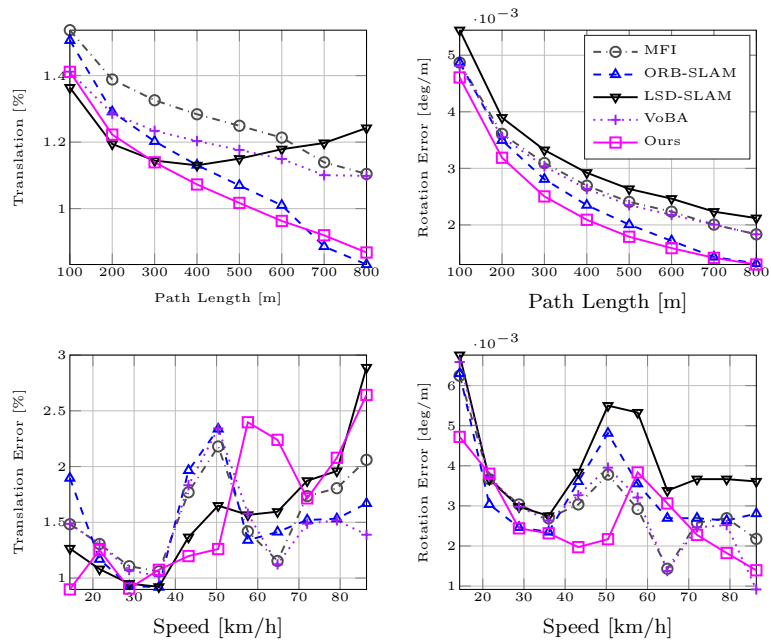

Performance of the algorithm is shown in \cref{fig:ba_results} and not only does it outperform the accuracy of (bundle adjusted and loop closed) ORB-SLAM, but also it outperforms other top performing algorithms, especially in the accuracy of estimating rotations. Compared algorithms include: ORB-SLAM~\cite{orbslam2015}, LSD-SLAM~\cite{engel2015_stereo_lsdslam,lsdslam}, VoBA~\cite{tardif2010new}, and MFI~\cite{badino2013visual}.

We note that sources of error in our algorithm are correlated with faster vehicle speeds. This is to be expected as the linearization of the photometric error holds only in a small neighborhood. This could be mitigated by implementing the algorithm in scale-space~\cite{lindeberg1993scale}, or improving the initialization quality of the scene structure (either by better stereo, or better scene points obtained from a geometric BA refinement step). Interestingly, however, the rotation error is reduced at high speeds which can be explained by lack of large rotations. The same behavior can be observed with LSD-SLAM's performance as both methods rely on the photometric error, but our rate of error reduction is higher due to the joint refinement of pose and structure parameters.

\begin{figure}
\centering
\setlength\fwidth{0.618\linewidth}
\setlength\fheight{\fwidth}
\input{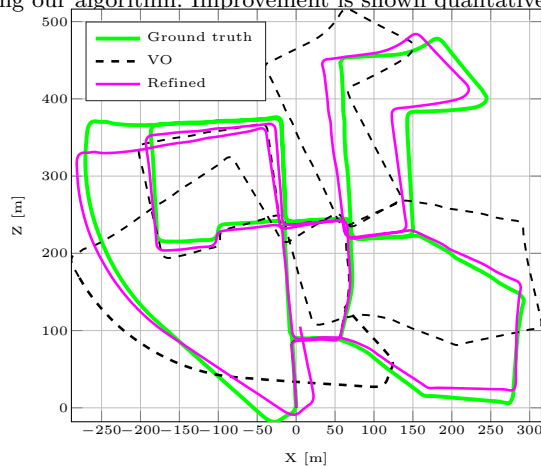}
\caption{{\small Magnitude of improvement starting from a poor initialization shown on the first sequence of the KITTI benchmark. Quantitative evaluation is shown in \cref{fig:ba_results}. We used a direct (correspondence-free) frame--frame VO method to initialize the pose parameters~\cite{bitplanes_vo}.}}
\label{fig:kitti_00_path}
\end{figure}

\subsubsection{Initializing with frame--frame VO} 
\begin{figure}
\centering
\setlength\fwidth{0.7\linewidth}
\setlength\fheight{2in}
\definecolor{mycolor1}{rgb}{1.00000,0.00000,1.00000}%
\begin{tikzpicture}

\begin{axis}[%
width=0.411\fwidth,
height=0.419\fheight,
at={(0\fwidth,0.581\fheight)},
scale only axis,
xmin=100,
xmax=800,
xlabel={{\tiny Path Length [m]}},
xmajorgrids,
ymin=2.81964854475823,
ymax=4.99481809372735,
ylabel={{\tiny Translation [\%]}},
ymajorgrids,
axis background/.style={fill=white},
name=plotname,
legend to name=kittifig,
legend columns = -1,
]
\addplot [color=black,dashed,line width=0.75pt,forget plot]
  table[row sep=crcr]{%
100	2.81964854475823\\
200	3.22639207629434\\
300	3.7202840499587\\
400	4.11275367486123\\
500	4.43299795076071\\
600	4.66553489152933\\
700	4.8443247844233\\
800	4.99481809372735\\
};
\addlegendentry{VO};
\addplot [color=mycolor1,solid,line width=0.75pt,forget plot]
  table[row sep=crcr]{%
100	2.8414749028795\\
200	2.85049448226185\\
300	2.97501121221901\\
400	3.03289330621795\\
500	3.0983612013709\\
600	3.13541968322386\\
700	3.16014837723682\\
800	3.17623130081643\\
};
\addlegendentry{Refined};
\addplot [color=black,dashed,line width=0.5pt,mark=o,mark options={solid},forget plot]
  table[row sep=crcr]{%
100	2.81964854475823\\
200	3.22639207629434\\
300	3.7202840499587\\
400	4.11275367486123\\
500	4.43299795076071\\
600	4.66553489152933\\
700	4.8443247844233\\
800	4.99481809372735\\
};
\addplot [color=mycolor1,solid,line width=0.5pt,mark=square,mark options={solid},forget plot]
  table[row sep=crcr]{%
100	2.8414749028795\\
200	2.85049448226185\\
300	2.97501121221901\\
400	3.03289330621795\\
500	3.0983612013709\\
600	3.13541968322386\\
700	3.16014837723682\\
800	3.17623130081643\\
};
\end{axis}
\begin{axis}[%
width=0.411\fwidth,
height=0.419\fheight,
at={(0.6\fwidth,0.581\fheight)},
scale only axis,
xmin=100,
xmax=800,
xlabel={{\tiny Path Length [m]}},
xmajorgrids,
ymin=0.00345901905644993,
ymax=0.0127523378658191,
ylabel={{\tiny Rotation Error [deg/m]}},
ymajorgrids,
axis background/.style={fill=white}
]
\addplot [color=black,dashed,line width=0.75pt,forget plot]
  table[row sep=crcr]{%
100	0.0127523378658191\\
200	0.0112526660971102\\
300	0.0105533664870707\\
400	0.0099616344828386\\
500	0.00965246639331242\\
600	0.00938783703131012\\
700	0.0091193084830129\\
800	0.00897650125378757\\
};

\addplot [color=mycolor1,solid,line width=0.75pt,forget plot]
  table[row sep=crcr]{%
100	0.00705583467513513\\
200	0.00548280314162604\\
300	0.00489261614009787\\
400	0.00432768920688974\\
500	0.00405513532944924\\
600	0.00374558288645794\\
700	0.00356788207343228\\
800	0.00345901905644993\\
};

\addplot [color=black,dashed,line width=0.5pt,mark=o,mark options={solid},forget plot]
  table[row sep=crcr]{%
100	0.0127523378658191\\
200	0.0112526660971102\\
300	0.0105533664870707\\
400	0.0099616344828386\\
500	0.00965246639331242\\
600	0.00938783703131012\\
700	0.0091193084830129\\
800	0.00897650125378757\\
};
\addplot [color=mycolor1,solid,line width=0.5pt,mark=square,mark options={solid},forget plot]
  table[row sep=crcr]{%
100	0.00705583467513513\\
200	0.00548280314162604\\
300	0.00489261614009787\\
400	0.00432768920688974\\
500	0.00405513532944924\\
600	0.00374558288645794\\
700	0.00356788207343228\\
800	0.00345901905644993\\
};
\end{axis}
\end{tikzpicture}%
%\ref{kittifig}
%\tikzexternaldisable\ref{kittifig}\tikzexternalenable
\caption{{\small Improvement in accuracy starting from a poor initialization using a frame--frame direct VO method with unbounded drift.}}
\label{fig:kitti_eval_ba}
\end{figure}
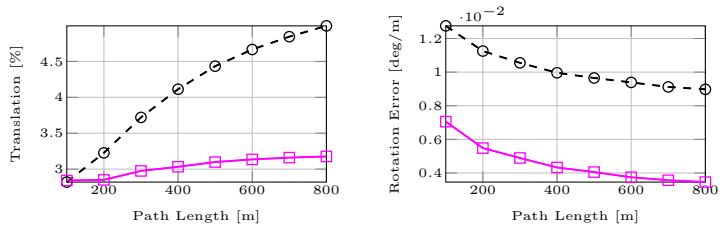

Surprisingly, and contrary to other image-based optimization schemes~\cite{furukawa2008accurate,delaunoy2014photometric}, our algorithm does not require an accurate initialization. \cref{fig:kitti_eval_ba} demonstrates a significant improvement in accuracy when the algorithm is initialized using frame--frame VO estimates with unbounded drift. Here, we used a direct method to initialize the camera pose without using any feature correspondences~\cite{bitplanes_vo}.

Interestingly, however, when starting from a poor initialization our algorithm does not attain the same accuracy as when initialized using a better quality starting point as shown in~\cref{fig:ba_results}. This leads us to conclude the algorithm is sensitive to the initialization conditions more so than traditional BA. Importantly, however, the algorithm is able to improve upon a poor initialization.

\subsubsection{Convergence characteristics and runtime}
As shown in \cref{fig:convergence} most of the photometric error is eliminated in the first five iterations of the minimization problem. While this is by no means a metric of quality, it is reassuring as it indicates a well-behaved optimization procedure.

After the first five iterations, the rate of the relative reduction in error slows down. This may be related to using linear interpolation to evaluate the photometric error, or the use of central differences to estimates gradients. Higher order interpolation methods~\cite{unser1999splines} or more accurate image gradients~\cite{farid2004differentiation} could have an influence on the rate of convergence and remain to be explored.

\begin{figure}
\centering
\setlength\fwidth{0.7\linewidth}
\setlength\fheight{1in}
\definecolor{mycolor1}{rgb}{0.00000,0.44700,0.74100}%
\definecolor{mycolor2}{rgb}{0.85000,0.32500,0.09800}%
\definecolor{mycolor3}{rgb}{0.92900,0.69400,0.12500}%
\definecolor{mycolor4}{rgb}{0.49400,0.18400,0.55600}%
\definecolor{mycolor5}{rgb}{0.46600,0.67400,0.18800}%
\definecolor{mycolor6}{rgb}{0.30100,0.74500,0.93300}%
\definecolor{mycolor7}{rgb}{0.63500,0.07800,0.18400}%
\begin{tikzpicture}

\begin{axis}[%
width=0.951\fwidth,
height=\fheight,
at={(0\fwidth,0\fheight)},
scale only axis,
xmin=0,
xmax=35,
xlabel={Iteration},
xmajorgrids,
ymin=-10,
ymax=400,
ylabel={Cost change},
ymajorgrids,
axis background/.style={fill=white},
axis x line*=bottom,
axis y line*=left
]
\addplot [color=mycolor1,solid,line width=1.0pt,forget plot]
  table[row sep=crcr]{%
1	386.798161703128\\
2	166.079817129532\\
3	77.5146542380116\\
4	31.103149294413\\
5	15.329656464231\\
6	3.73814934097709\\
7	2.87931931247931\\
8	2.81581761570942\\
9	9.97794176292427\\
10	4.69971041636381\\
11	0.231825188328003\\
12	1.5788581734173\\
13	0.0475767226876087\\
14	0.997353111758457\\
15	0.519898194987945\\
16	0.252342477073398\\
17	0.173942515277304\\
18	0.0852881004352639\\
19	0.0127183699537454\\
20	0.0579324500472467\\
21	0.0706530172742532\\
22	0.0542480672606871\\
23	0.0583265868954186\\
24	0.0575375382659331\\
25	0.0484213386885131\\
26	0.0622287836513351\\
27	0.0657761958123046\\
28	0.0611666490158314\\
29	0.0562612027692921\\
30	0.0409832441328035\\
31	0.033037672719729\\
32	0.0332871267783048\\
33	0.0197329663824348\\
};
\addplot [color=mycolor2,solid,forget plot]
  table[row sep=crcr]{%
1	164.46381450607\\
2	65.8159173928876\\
3	24.7511975999066\\
4	13.0483465753454\\
5	9.53670351734979\\
6	8.89898529426296\\
7	2.31901100673667\\
8	1.61565976091026\\
9	3.27468809701702\\
10	0.709488235506797\\
11	10.0491753511475\\
12	0.586135043573393\\
13	3.31144413815525\\
14	3.0556748192065\\
15	0.439373082279644\\
16	1.77372470008987\\
17	1.25633223792852\\
18	0.107153394422255\\
19	0.545062613461596\\
20	0.26385042148911\\
21	0.477198620644231\\
22	0.122472183059017\\
23	0.686628604983298\\
24	0.184663855200597\\
25	0.152595879377259\\
26	0.122469828144858\\
27	0.206104957062223\\
28	0.0263379645797954\\
29	0.0429588267961662\\
30	0.00853607201486284\\
};
\addplot [color=mycolor3,solid,forget plot]
  table[row sep=crcr]{%
1	54.6091818850321\\
2	26.2846324200598\\
3	1.71648762035295\\
4	11.1277793188204\\
5	6.07568588113429\\
6	3.43915630405513\\
7	1.20972385352434\\
8	9.8909284230989\\
9	3.2180267874254\\
10	0.406995418716065\\
11	3.72365026534294\\
12	2.00104963838396\\
13	2.66652746863406\\
14	3.73829477175434\\
15	1.04784451829255\\
16	0.36213355531936\\
17	2.81578440589283\\
18	0.936352270074622\\
19	0.40384632190387\\
20	0.893315663381486\\
21	2.12027194479515\\
22	0.450088672666197\\
23	0.780163464696329\\
24	0.124783491011215\\
25	0.0824320893527783\\
26	0.0275056769280582\\
};
\addplot [color=mycolor4,solid,forget plot]
  table[row sep=crcr]{%
1	161.017168836027\\
2	29.741046845973\\
3	9.89958634275035\\
4	12.989072916379\\
5	1.63640988684256\\
6	2.81052931910222\\
7	5.5836333119239\\
8	0.525718937295551\\
9	0.25391842441968\\
10	5.02300392035568\\
11	0.11763044151985\\
12	2.11775945673276\\
13	0.690925956624596\\
14	8.41027449677722\\
15	3.0494164836723\\
16	4.77436370185796\\
17	5.63610070871869\\
18	4.15401002407111\\
19	4.41518211235461\\
20	1.54162395264962\\
21	0.751774398241878\\
22	2.8453648356849\\
23	1.18162581722345\\
24	0.794446844166941\\
25	0.28566159926163\\
26	2.00847489139642\\
27	0.322974640324219\\
28	0.315465383731635\\
29	0.110094735656503\\
30	0.0915389141277956\\
31	0.0442930256506315\\
32	0.00750881381782165\\
};
\addplot [color=mycolor5,solid,forget plot]
  table[row sep=crcr]{%
1	135.880322361485\\
2	35.5009840057146\\
3	7.57464682937916\\
4	6.55454061160094\\
5	12.7818909725891\\
6	2.51588368881789\\
7	6.30900714594759\\
8	0.437149275978754\\
9	0.810859292047326\\
10	2.75558016112882\\
11	0.753015778307145\\
12	2.99750883802039\\
13	3.00451461639523\\
14	0.445439284589156\\
15	1.23474655458608\\
16	8.06553989112626\\
17	3.35219143707468\\
18	1.04310587223426\\
19	0.725931388431718\\
20	9.81690426844762\\
21	0.208001686421994\\
22	6.21610699046278\\
23	1.95263213347562\\
24	1.46813601804547\\
25	2.89909902849308\\
26	0.116101276959625\\
27	0.583563307129225\\
28	0.189229341777946\\
29	0.24369799540159\\
};
\addplot [color=mycolor6,solid,forget plot]
  table[row sep=crcr]{%
1	70.154377106488\\
2	12.5300989101152\\
3	1.108975793637\\
4	0.969363263036485\\
5	11.5732605671174\\
6	0.482775281871\\
7	2.90371865088809\\
8	1.70207576680241\\
9	2.90864071398209\\
10	3.87097271120638\\
11	0.546707257964954\\
12	3.40826141919888\\
13	2.46568413348587\\
14	2.82551917476167\\
15	1.57767005391906\\
16	3.46487527017052\\
17	0.254030039158806\\
18	2.92320084612265\\
19	2.10150734240824\\
20	1.04870033351926\\
21	0.646022690229529\\
22	0.0453713755127865\\
23	0.367152630181408\\
24	0.0254223551710311\\
25	0.0116567907780336\\
};
\addplot [color=mycolor7,solid,forget plot]
  table[row sep=crcr]{%
1	130.701503081518\\
2	34.6076927235454\\
3	20.2429778345672\\
4	1.45639299428331\\
5	12.8038683112668\\
6	0.662285139873347\\
7	9.28622934145642\\
8	2.98339924790707\\
9	9.73227535767001\\
10	3.16183702432136\\
11	5.98366093350114\\
12	0.106741166547636\\
13	11.7660709930606\\
14	0.752324926000711\\
15	14.1727254705534\\
16	2.64314474318144\\
17	6.69421791777404\\
18	2.67566478165827\\
19	6.69355475448538\\
20	1.07808502386024\\
21	5.14407424938327\\
22	2.00509918805119\\
23	0.92820870699461\\
24	1.0519351806106\\
25	1.5223242156344\\
26	0.967781465585176\\
27	0.547282196222113\\
28	0.0213748080145706\\
29	0.247627306397135\\
30	0.0461055794689855\\
};
\addplot [color=mycolor1,solid,forget plot]
  table[row sep=crcr]{%
1	60.3802828644252\\
2	0.802446748618422\\
3	8.42178592520486\\
4	2.60724912621117\\
5	3.84987493627614\\
6	0.596474248485265\\
7	0.18980477922878\\
8	0.179611788867078\\
9	4.11827903180756\\
10	2.07189117114194\\
11	3.25479179848435\\
12	4.51524724678438\\
13	4.24402729366966\\
14	1.03403729798993\\
15	5.68188019564786\\
16	1.02471008279281\\
17	4.35952260209979\\
18	1.79128710780924\\
19	1.77859906805634\\
20	0.383530288503607\\
21	2.07034337978439\\
22	1.04148325148299\\
23	1.45492634984794\\
24	0.0970108172259643\\
25	0.776107063128165\\
26	0.322484017779516\\
27	0.00296044535298279\\
28	0.0185283791452093\\
29	0.00612315892976767\\
30	0.0133193761412258\\
};
\addplot [color=mycolor2,solid,forget plot]
  table[row sep=crcr]{%
1	83.0816502053012\\
2	2.73679614742241\\
3	13.3680955940804\\
4	1.33409083772767\\
5	2.06941242396351\\
6	0.125903066207684\\
7	9.99584464158579\\
8	5.9696593607855\\
9	1.81517706019531\\
10	2.52657684922224\\
11	3.76743926761583\\
12	1.93701218217257\\
13	2.18615171005013\\
14	3.22426263133457\\
15	2.0147340593976\\
16	0.774146606839167\\
17	2.67530149089998\\
18	0.352398084964079\\
19	1.89448172137691\\
20	0.106186395575151\\
21	0.673999701495177\\
22	0.345561433276089\\
23	0.0756586721454369\\
24	0.0737491749628134\\
25	0.00352914268660243\\
};
\addplot [color=mycolor3,solid,forget plot]
  table[row sep=crcr]{%
1	151.55800813364\\
2	30.2473825698953\\
3	17.1206143265886\\
4	13.9364193443666\\
5	7.09109779434357\\
6	1.9726262629415\\
7	6.29396397766868\\
8	0.722988539357175\\
9	1.30598073467445\\
10	1.76105302529186\\
11	4.31319756368657\\
12	2.88786225558306\\
13	4.58392559086269\\
14	1.38728670060482\\
15	8.2836624699612\\
16	3.27130948533113\\
17	5.44429499733678\\
18	4.50078567268088\\
19	4.21563556883962\\
20	0.276417451498219\\
21	3.07343057363664\\
22	1.28778545419459\\
23	0.429147629706222\\
24	0.0822648519999802\\
25	0.311642102233463\\
};
\end{axis}
\end{tikzpicture}%
\caption{{\small Rate of error reduction at every iteration shown for the first $10$ sliding windows, each with $5$ frames. The thicker line shows the first bundle, which has the highest error. Most of the error is eliminated with the first $5$ iterations.}}
\label{fig:convergence}
\end{figure}
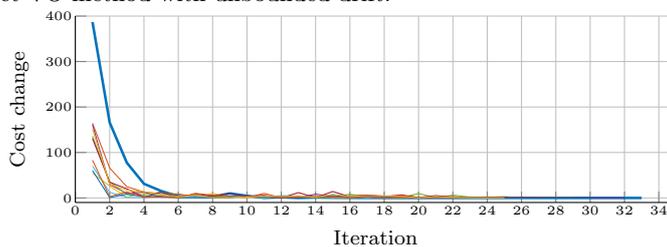

\begin{figure}
\centering
\begin{minipage}{0.5\linewidth}
\setlength\fwidth{0.9\linewidth}
\setlength\fheight{1in}
\definecolor{mycolor1}{rgb}{0.83137,0.81569,0.78431}%
\begin{tikzpicture}

\begin{axis}[%
width=0.951\fwidth,
height=\fheight,
at={(0\fwidth,0\fheight)},
scale only axis,
point meta min=1,
point meta max=2,
colormap={mymap}{[1pt] rgb(0pt)=(0.2081,0.1663,0.5292); rgb(1pt)=(0.211624,0.189781,0.577676); rgb(2pt)=(0.212252,0.213771,0.626971); rgb(3pt)=(0.2081,0.2386,0.677086); rgb(4pt)=(0.195905,0.264457,0.7279); rgb(5pt)=(0.170729,0.291938,0.779248); rgb(6pt)=(0.125271,0.324243,0.830271); rgb(7pt)=(0.0591333,0.359833,0.868333); rgb(8pt)=(0.0116952,0.38751,0.881957); rgb(9pt)=(0.00595714,0.408614,0.882843); rgb(10pt)=(0.0165143,0.4266,0.878633); rgb(11pt)=(0.0328524,0.443043,0.871957); rgb(12pt)=(0.0498143,0.458571,0.864057); rgb(13pt)=(0.0629333,0.47369,0.855438); rgb(14pt)=(0.0722667,0.488667,0.8467); rgb(15pt)=(0.0779429,0.503986,0.838371); rgb(16pt)=(0.0793476,0.520024,0.831181); rgb(17pt)=(0.0749429,0.537543,0.826271); rgb(18pt)=(0.0640571,0.556986,0.823957); rgb(19pt)=(0.0487714,0.577224,0.822829); rgb(20pt)=(0.0343429,0.596581,0.819852); rgb(21pt)=(0.0265,0.6137,0.8135); rgb(22pt)=(0.0238905,0.628662,0.803762); rgb(23pt)=(0.0230905,0.641786,0.791267); rgb(24pt)=(0.0227714,0.653486,0.776757); rgb(25pt)=(0.0266619,0.664195,0.760719); rgb(26pt)=(0.0383714,0.674271,0.743552); rgb(27pt)=(0.0589714,0.683757,0.725386); rgb(28pt)=(0.0843,0.692833,0.706167); rgb(29pt)=(0.113295,0.7015,0.685857); rgb(30pt)=(0.145271,0.709757,0.664629); rgb(31pt)=(0.180133,0.717657,0.642433); rgb(32pt)=(0.217829,0.725043,0.619262); rgb(33pt)=(0.258643,0.731714,0.595429); rgb(34pt)=(0.302171,0.737605,0.571186); rgb(35pt)=(0.348167,0.742433,0.547267); rgb(36pt)=(0.395257,0.7459,0.524443); rgb(37pt)=(0.44201,0.748081,0.503314); rgb(38pt)=(0.487124,0.749062,0.483976); rgb(39pt)=(0.530029,0.749114,0.466114); rgb(40pt)=(0.570857,0.748519,0.44939); rgb(41pt)=(0.609852,0.747314,0.433686); rgb(42pt)=(0.6473,0.7456,0.4188); rgb(43pt)=(0.683419,0.743476,0.404433); rgb(44pt)=(0.71841,0.741133,0.390476); rgb(45pt)=(0.752486,0.7384,0.376814); rgb(46pt)=(0.785843,0.735567,0.363271); rgb(47pt)=(0.818505,0.732733,0.34979); rgb(48pt)=(0.850657,0.7299,0.336029); rgb(49pt)=(0.882433,0.727433,0.3217); rgb(50pt)=(0.913933,0.725786,0.306276); rgb(51pt)=(0.944957,0.726114,0.288643); rgb(52pt)=(0.973895,0.731395,0.266648); rgb(53pt)=(0.993771,0.745457,0.240348); rgb(54pt)=(0.999043,0.765314,0.216414); rgb(55pt)=(0.995533,0.786057,0.196652); rgb(56pt)=(0.988,0.8066,0.179367); rgb(57pt)=(0.978857,0.827143,0.163314); rgb(58pt)=(0.9697,0.848138,0.147452); rgb(59pt)=(0.962586,0.870514,0.1309); rgb(60pt)=(0.958871,0.8949,0.113243); rgb(61pt)=(0.959824,0.921833,0.0948381); rgb(62pt)=(0.9661,0.951443,0.0755333); rgb(63pt)=(0.9763,0.9831,0.0538)},
xmin=3,
xmax=73,
xlabel={{\small Number of iterations}},
ymin=0,
ymax=1243,
axis background/.style={fill=white}
]

\addplot[area legend,solid,table/row sep=crcr,patch,patch type=rectangle,fill=mycolor1,faceted color=white!15!black,forget plot,patch table with point meta={%
1	2	3	4	1\\
6	7	8	9	1\\
11	12	13	14	1\\
16	17	18	19	1\\
21	22	23	24	1\\
26	27	28	29	1\\
31	32	33	34	1\\
36	37	38	39	1\\
41	42	43	44	1\\
46	47	48	49	1\\
51	52	53	54	1\\
56	57	58	59	1\\
61	62	63	64	1\\
66	67	68	69	1\\
71	72	73	74	1\\
76	77	78	79	1\\
81	82	83	84	1\\
86	87	88	89	1\\
91	92	93	94	1\\
96	97	98	99	1\\
101	102	103	104	1\\
106	107	108	109	1\\
111	112	113	114	1\\
116	117	118	119	1\\
121	122	123	124	1\\
}]
table[row sep=crcr] {%
x	y\\
3	0\\
3	0\\
3	4\\
5.8	4\\
5.8	0\\
5.8	0\\
5.8	0\\
5.8	12\\
8.6	12\\
8.6	0\\
8.6	0\\
8.6	0\\
8.6	11\\
11.4	11\\
11.4	0\\
11.4	0\\
11.4	0\\
11.4	20\\
14.2	20\\
14.2	0\\
14.2	0\\
14.2	0\\
14.2	20\\
17	20\\
17	0\\
17	0\\
17	0\\
17	16\\
19.8	16\\
19.8	0\\
19.8	0\\
19.8	0\\
19.8	28\\
22.6	28\\
22.6	0\\
22.6	0\\
22.6	0\\
22.6	120\\
25.4	120\\
25.4	0\\
25.4	0\\
25.4	0\\
25.4	252\\
28.2	252\\
28.2	0\\
28.2	0\\
28.2	0\\
28.2	710\\
31	710\\
31	0\\
31	0\\
31	0\\
31	793\\
33.8	793\\
33.8	0\\
33.8	0\\
33.8	0\\
33.8	1243\\
36.6	1243\\
36.6	0\\
36.6	0\\
36.6	0\\
36.6	749\\
39.4	749\\
39.4	0\\
39.4	0\\
39.4	0\\
39.4	302\\
42.2	302\\
42.2	0\\
42.2	0\\
42.2	0\\
42.2	106\\
45	106\\
45	0\\
45	0\\
45	0\\
45	54\\
47.8	54\\
47.8	0\\
47.8	0\\
47.8	0\\
47.8	40\\
50.6	40\\
50.6	0\\
50.6	0\\
50.6	0\\
50.6	23\\
53.4	23\\
53.4	0\\
53.4	0\\
53.4	0\\
53.4	11\\
56.2	11\\
56.2	0\\
56.2	0\\
56.2	0\\
56.2	8\\
59	8\\
59	0\\
59	0\\
59	0\\
59	4\\
61.8	4\\
61.8	0\\
61.8	0\\
61.8	0\\
61.8	6\\
64.6	6\\
64.6	0\\
64.6	0\\
64.6	0\\
64.6	2\\
67.4	2\\
67.4	0\\
67.4	0\\
67.4	0\\
67.4	2\\
70.2	2\\
70.2	0\\
70.2	0\\
70.2	0\\
70.2	1\\
73	1\\
73	0\\
73	0\\
};
\end{axis}
\end{tikzpicture}%
\label{fig:num_iters_hist}
\end{minipage}%
\begin{minipage}{0.5\linewidth}
\setlength\fwidth{0.9\linewidth}
\setlength\fheight{1in}
\definecolor{mycolor1}{rgb}{0.83137,0.81569,0.78431}%
\begin{tikzpicture}
\begin{axis}[%
width=0.951\fwidth,
height=\fheight,
at={(0\fwidth,0\fheight)},
scale only axis,
point meta min=1,
point meta max=2,
colormap={mymap}{[1pt] rgb(0pt)=(0.2081,0.1663,0.5292); rgb(1pt)=(0.211624,0.189781,0.577676); rgb(2pt)=(0.212252,0.213771,0.626971); rgb(3pt)=(0.2081,0.2386,0.677086); rgb(4pt)=(0.195905,0.264457,0.7279); rgb(5pt)=(0.170729,0.291938,0.779248); rgb(6pt)=(0.125271,0.324243,0.830271); rgb(7pt)=(0.0591333,0.359833,0.868333); rgb(8pt)=(0.0116952,0.38751,0.881957); rgb(9pt)=(0.00595714,0.408614,0.882843); rgb(10pt)=(0.0165143,0.4266,0.878633); rgb(11pt)=(0.0328524,0.443043,0.871957); rgb(12pt)=(0.0498143,0.458571,0.864057); rgb(13pt)=(0.0629333,0.47369,0.855438); rgb(14pt)=(0.0722667,0.488667,0.8467); rgb(15pt)=(0.0779429,0.503986,0.838371); rgb(16pt)=(0.0793476,0.520024,0.831181); rgb(17pt)=(0.0749429,0.537543,0.826271); rgb(18pt)=(0.0640571,0.556986,0.823957); rgb(19pt)=(0.0487714,0.577224,0.822829); rgb(20pt)=(0.0343429,0.596581,0.819852); rgb(21pt)=(0.0265,0.6137,0.8135); rgb(22pt)=(0.0238905,0.628662,0.803762); rgb(23pt)=(0.0230905,0.641786,0.791267); rgb(24pt)=(0.0227714,0.653486,0.776757); rgb(25pt)=(0.0266619,0.664195,0.760719); rgb(26pt)=(0.0383714,0.674271,0.743552); rgb(27pt)=(0.0589714,0.683757,0.725386); rgb(28pt)=(0.0843,0.692833,0.706167); rgb(29pt)=(0.113295,0.7015,0.685857); rgb(30pt)=(0.145271,0.709757,0.664629); rgb(31pt)=(0.180133,0.717657,0.642433); rgb(32pt)=(0.217829,0.725043,0.619262); rgb(33pt)=(0.258643,0.731714,0.595429); rgb(34pt)=(0.302171,0.737605,0.571186); rgb(35pt)=(0.348167,0.742433,0.547267); rgb(36pt)=(0.395257,0.7459,0.524443); rgb(37pt)=(0.44201,0.748081,0.503314); rgb(38pt)=(0.487124,0.749062,0.483976); rgb(39pt)=(0.530029,0.749114,0.466114); rgb(40pt)=(0.570857,0.748519,0.44939); rgb(41pt)=(0.609852,0.747314,0.433686); rgb(42pt)=(0.6473,0.7456,0.4188); rgb(43pt)=(0.683419,0.743476,0.404433); rgb(44pt)=(0.71841,0.741133,0.390476); rgb(45pt)=(0.752486,0.7384,0.376814); rgb(46pt)=(0.785843,0.735567,0.363271); rgb(47pt)=(0.818505,0.732733,0.34979); rgb(48pt)=(0.850657,0.7299,0.336029); rgb(49pt)=(0.882433,0.727433,0.3217); rgb(50pt)=(0.913933,0.725786,0.306276); rgb(51pt)=(0.944957,0.726114,0.288643); rgb(52pt)=(0.973895,0.731395,0.266648); rgb(53pt)=(0.993771,0.745457,0.240348); rgb(54pt)=(0.999043,0.765314,0.216414); rgb(55pt)=(0.995533,0.786057,0.196652); rgb(56pt)=(0.988,0.8066,0.179367); rgb(57pt)=(0.978857,0.827143,0.163314); rgb(58pt)=(0.9697,0.848138,0.147452); rgb(59pt)=(0.962586,0.870514,0.1309); rgb(60pt)=(0.958871,0.8949,0.113243); rgb(61pt)=(0.959824,0.921833,0.0948381); rgb(62pt)=(0.9661,0.951443,0.0755333); rgb(63pt)=(0.9763,0.9831,0.0538)},
xmin=0,
xmax=7,
xlabel={{\small Time in seconds}},
ymin=0,
ymax=800,
axis background/.style={fill=white}
]

\addplot[area legend,solid,table/row sep=crcr,patch,patch type=rectangle,fill=mycolor1,faceted color=white!15!black,forget plot,patch table with point meta={%
1	2	3	4	1\\
6	7	8	9	1\\
11	12	13	14	1\\
16	17	18	19	1\\
21	22	23	24	1\\
26	27	28	29	1\\
31	32	33	34	1\\
36	37	38	39	1\\
41	42	43	44	1\\
46	47	48	49	1\\
51	52	53	54	1\\
56	57	58	59	1\\
61	62	63	64	1\\
66	67	68	69	1\\
71	72	73	74	1\\
76	77	78	79	1\\
81	82	83	84	1\\
86	87	88	89	1\\
91	92	93	94	1\\
96	97	98	99	1\\
101	102	103	104	1\\
106	107	108	109	1\\
111	112	113	114	1\\
116	117	118	119	1\\
121	122	123	124	1\\
}]
table[row sep=crcr] {%
x	y\\
0.143433912016917	0\\
0.143433912016917	0\\
0.143433912016917	8\\
0.400369384137448	8\\
0.400369384137448	0\\
0.400369384137448	0\\
0.400369384137448	0\\
0.400369384137448	19\\
0.657304856257979	19\\
0.657304856257979	0\\
0.657304856257979	0\\
0.657304856257979	0\\
0.657304856257979	108\\
0.91424032837851	108\\
0.91424032837851	0\\
0.91424032837851	0\\
0.91424032837851	0\\
0.91424032837851	280\\
1.17117580049904	280\\
1.17117580049904	0\\
1.17117580049904	0\\
1.17117580049904	0\\
1.17117580049904	493\\
1.42811127261957	493\\
1.42811127261957	0\\
1.42811127261957	0\\
1.42811127261957	0\\
1.42811127261957	699\\
1.6850467447401	699\\
1.6850467447401	0\\
1.6850467447401	0\\
1.6850467447401	0\\
1.6850467447401	784\\
1.94198221686063	784\\
1.94198221686063	0\\
1.94198221686063	0\\
1.94198221686063	0\\
1.94198221686063	689\\
2.19891768898116	689\\
2.19891768898116	0\\
2.19891768898116	0\\
2.19891768898116	0\\
2.19891768898116	479\\
2.4558531611017	479\\
2.4558531611017	0\\
2.45585316110169	0\\
2.45585316110169	0\\
2.45585316110169	366\\
2.71278863322223	366\\
2.71278863322223	0\\
2.71278863322223	0\\
2.71278863322223	0\\
2.71278863322223	230\\
2.96972410534276	230\\
2.96972410534276	0\\
2.96972410534276	0\\
2.96972410534276	0\\
2.96972410534276	152\\
3.22665957746329	152\\
3.22665957746329	0\\
3.22665957746329	0\\
3.22665957746329	0\\
3.22665957746329	88\\
3.48359504958382	88\\
3.48359504958382	0\\
3.48359504958382	0\\
3.48359504958382	0\\
3.48359504958382	50\\
3.74053052170435	50\\
3.74053052170435	0\\
3.74053052170435	0\\
3.74053052170435	0\\
3.74053052170435	41\\
3.99746599382488	41\\
3.99746599382488	0\\
3.99746599382488	0\\
3.99746599382488	0\\
3.99746599382488	24\\
4.25440146594541	24\\
4.25440146594541	0\\
4.25440146594541	0\\
4.25440146594541	0\\
4.25440146594541	8\\
4.51133693806594	8\\
4.51133693806594	0\\
4.51133693806594	0\\
4.51133693806594	0\\
4.51133693806594	9\\
4.76827241018647	9\\
4.76827241018647	0\\
4.76827241018647	0\\
4.76827241018647	0\\
4.76827241018647	3\\
5.025207882307	3\\
5.025207882307	0\\
5.025207882307	0\\
5.025207882307	0\\
5.025207882307	2\\
5.28214335442753	2\\
5.28214335442753	0\\
5.28214335442753	0\\
5.28214335442753	0\\
5.28214335442753	2\\
5.53907882654807	2\\
5.53907882654807	0\\
5.53907882654807	0\\
5.53907882654807	0\\
5.53907882654807	0\\
5.7960142986686	0\\
5.7960142986686	0\\
5.7960142986686	0\\
5.7960142986686	0\\
5.7960142986686	1\\
6.05294977078913	1\\
6.05294977078913	0\\
6.05294977078913	0\\
6.05294977078913	0\\
6.05294977078913	0\\
6.30988524290966	0\\
6.30988524290966	0\\
6.30988524290966	0\\
6.30988524290966	0\\
6.30988524290966	2\\
6.56682071503019	2\\
6.56682071503019	0\\
6.56682071503019	0\\
};
\end{axis}
\end{tikzpicture}%
\label{fig:time_in_sec}
\end{minipage}%
\caption{{\small Histogram of the number of iterations (on the left) and runtime (on the right). The median number of iterations is $34$, with a standard deviation of $6.02$. The median run time is $1.89$, mean $1.98$ and standard deviation of $0.69$. The runtime is reported for sliding window of $5$ frames on the KITTI benchmark.}}
\label{fig:timing}
\end{figure}
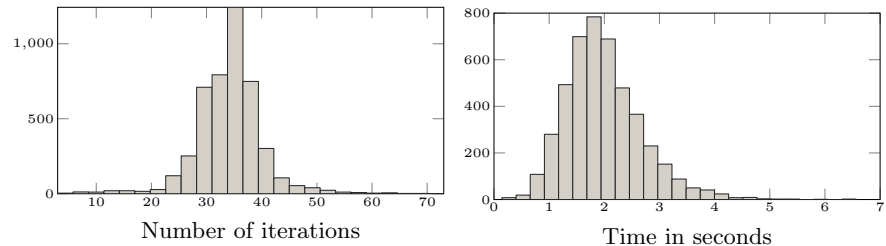

The number of iterations and cumulative runtime per sliding window of $5$ frames is shown in~\cref{fig:timing}. The median number of iterations is $34$ with a standard deviation of $\approx 6$. Statistics are computed on the KITTI dataset frames. The runtime is $\approx 2s$ per sliding window ($\SI{400}{\milli\second}$ per frame) using a laptop with a dual core processor clocked at \SI{2.8}{\GHz} and \SI{8}{\giga\byte} of RAM, which limits parallelism. We note that it is possible to improve the runtime of the proposed method significantly using the CPU, or the GPU. The bottleneck of the proposed algorithm is image interpolation (which can be done efficiently with SIMD instructions) and the reliance on automatic differentiation (which limits any code optimization as the code must remain simple for automatic differentiation to work).

\subsection{The M{\'{a}}laga Stereo Dataset}
The M{\'{a}}laga dataset~\cite{badino2013visual} is a particularly challenging dataset for VSLAM. The dataset features driving in urban areas using a small baseline stereo camera at resolution $800\times 600$. The baseline of the stereo is \SI{12}{\centi\metre} which provides little parallax for resolving distal observations. In addition, the camera is pointed upward towards the sky to avoid imaging the vehicle, which limits using points on the ground plane and closer to the camera. We use extracts $1$, $3$, and $6$ in our evaluation.

\begin{figure}
\centering
\includegraphics[width=0.65\linewidth]{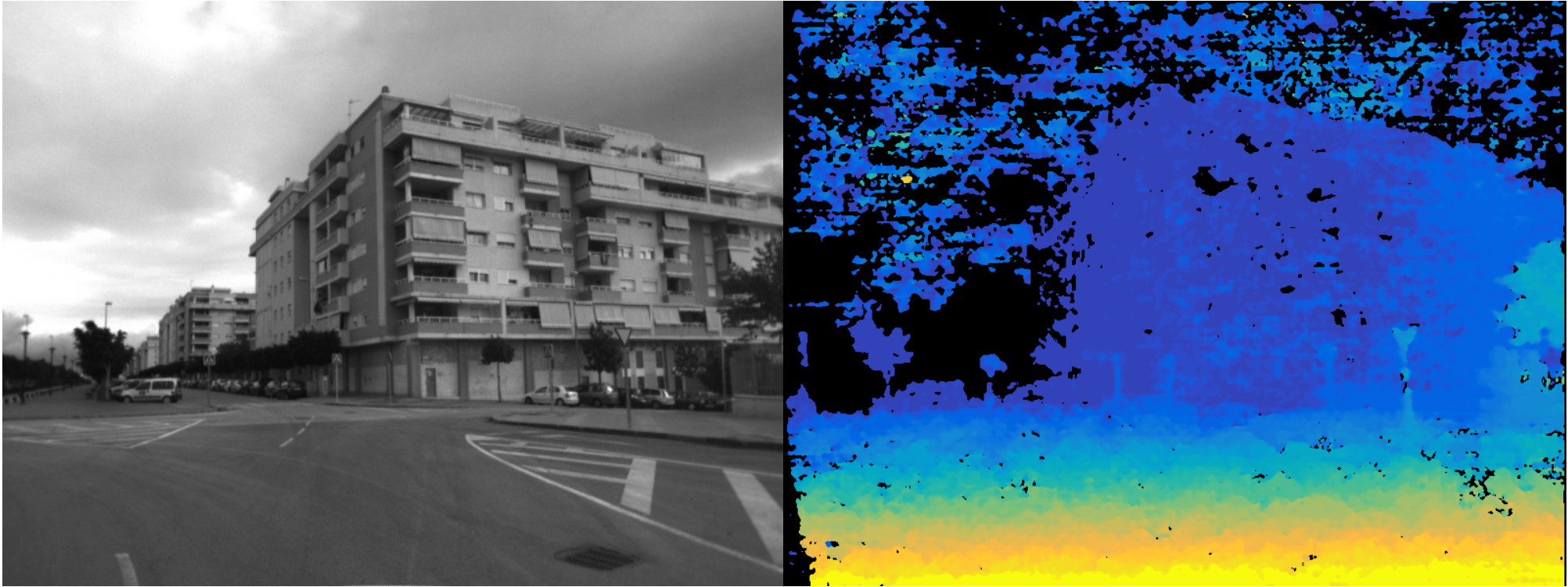}
\caption{{\small Quality of stereo use to initialize our algorithm on the Malaga dataset. The pixels marked in black indicate missing disparity estimates.}}
\label{fig:malaga_stereo}
\end{figure}

Our experimental setup is similar to the KITTI dataset. However, we estimate the stereo using the SGM algorithm~\cite{sgm}, as implemented in the OpenCV library. The stereo is used to estimate $16$ disparities with a SAD block size of $5\times 5$. The quality of stereo is low due to the difficulty of the dataset as shown in~\cref{fig:malaga_stereo}. We did not observe a significant difference in performance when using block matching instead of SGM.

\begin{figure}
\centering
\includegraphics[width=0.9\linewidth,trim={0 3cm 0 0}]{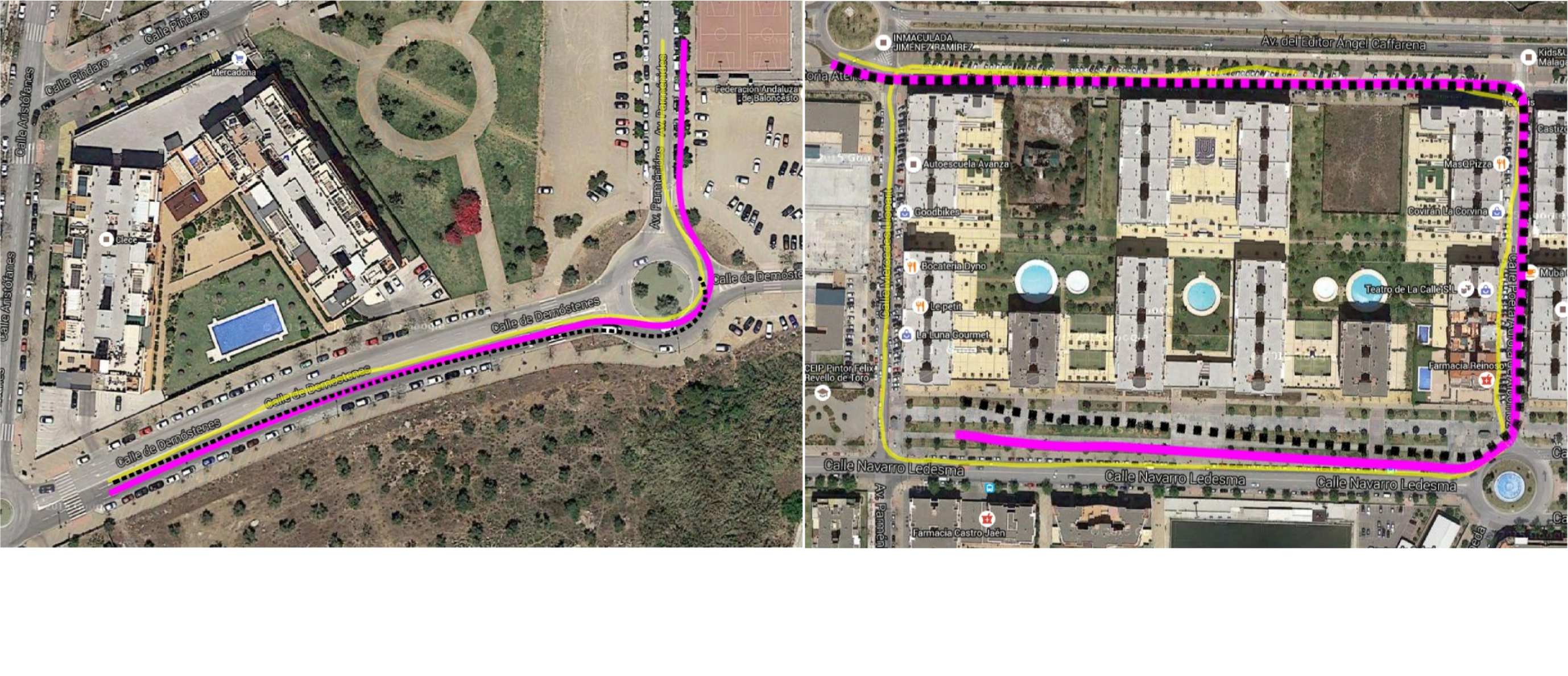}
\caption{{\small Our algorithm (magenta) compared with ORB-SLAM (dashed) against GPS (yellow) on extracts $3$ and $6$ of the Malaga dataset. For extract $3$ ORB-SLAM loses tracking during the roundabout, where our algorithm continues without an initialization. Results for extract $6$ are shown up to frame $3000$ as ORB-SLAM loses tracking. The figure is best viewed in color. (Maps courtesy of Google Maps.)}}
\label{fig:malaga_paths}
\end{figure}

The Malaga dataset provides GPS measurements, but they are not accurate enough for quantitative evaluation. The GPS path, however, is sufficient to qualitatively demonstrate precision. Results are shown in \cref{fig:malaga_paths} in comparison with ORB-SLAM~\cite{orbslam2015}, which we used its pose output to initialize our algorithm. We note that in extract $3$ of the Malaga dataset (shown on the left in~\cref{fig:malaga_paths}), ORB-SLAM loses tracking during the turn and our algorithm continues \emph{without} initialization.

\begin{figure}
\centering
\includegraphics[width=0.9\linewidth]{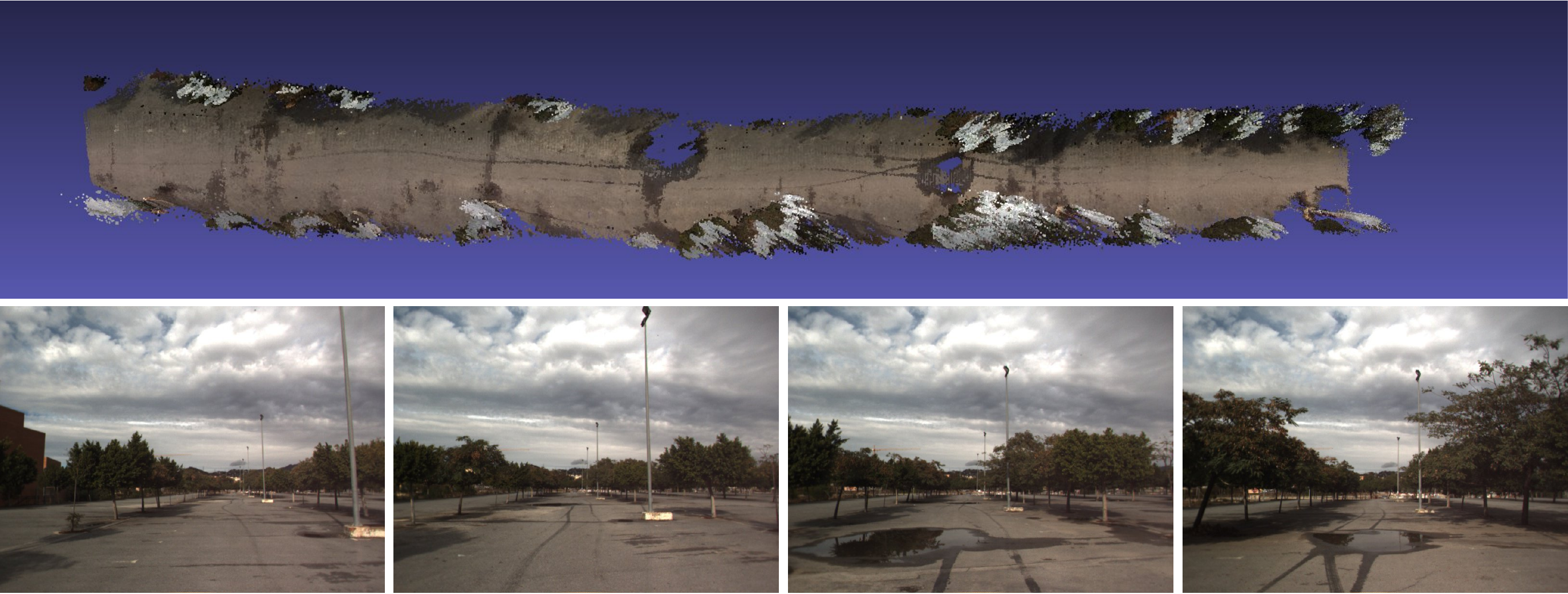}
\caption{Dense map from Malaga dataset extract $1$. The map is computed by stitching together SGM disparity with the refined camera pose.}
\label{fig:malaga_01_map}
\end{figure}

To assess the quality of pose estimates, we demonstrate results on a dense reconstruction procedure shown in \cref{fig:malaga_01_map}. Using the estimated camera trajectory, we chain the first $\SI{6}{\metre}$ of the disparity estimates to generate a dense map. As shown in~\cref{fig:malaga_01_map}, the quality of pose estimates appears to be good.

\begin{comment}
It is also instructive to observe errors in terms of a dense reconstruction of the environment. To this end, we implemented a simple dense reconstruction system to illustrate the accuracy of pose estimation. For every frame, we compute a dense stereo image up to a few meters in front of the car using SGM and stitch those maps using pose estimates. Results are shown in~\cref{fig:3dmaps}.

\begin{figure}
\caption{{\small 3D reconstruction results from stereo aimed at evaluating pose accuracy.}}
\label{fig:3dmaps}
\end{figure}
\end{comment}

\section{Related Work}
\subsubsection{Geometric BA}
BA has a long and rich history in computer vision, photogrammetry and robotics~\cite{ba}. BA is a large geometric minimization problem with the important property that variable interactions result in a \emph{sparse} system of linear equations. This sparsity is key to enabling large--scale applications~\cite{bal2010,konolige2010sparse}. Exploiting this sparsity is also key to obtaining precise results efficiently~\cite{Jeong2010,Engels06bundleadjustment}. The efficiency of BA has been an important research topic especially when handling large datasets~\cite{wu2011multicore,Ni2007} and in robotics applications~\cite{Konolige2008,Kaess2011,Kaess08tro}. Optimality and convergence properties of BA have been studied at length~\cite{Kahl2008,hartley13,Hartley2006} and remain of interest to date~\cite{Aftab2015}. All the aforementioned research in geometric BA could be integrated into the proposed photometric BA framework.

\subsubsection{Direct multi-frame alignment}
By direct alignment we mean algorithms that estimate the parameters of interest from the image data directly and without relying on sparse features as an intermediate representation of the image~\cite{irani1999direct}. The fundamental differences between direct methods (like the one proposed herein) and the commonly used feature-based pipeline is how the correspondence problem is tackled and is not related to the density of the reconstruction.

In the feature-based pipeline~\cite{torr00}, structure and motion parameters are estimated from known, pre-computed and fixed correspondence. In contrast, the direct pipeline to motion estimation does not used fixed correspondences. Instead, the correspondences are estimated as a byproduct of directly estimating the parameters of interest (\eg structure and motion).

The use of direct algorithms for SFM applications was studied for small--scale problems~\cite{Horn1988,Oliensis2000,Mandelbaum99,SteinShashua2000}, but feature-based alignment has proven more successful in handling wide baseline matching problems~\cite{torr00} as small pixel displacements is an integral assumption for direct methods. Nonetheless, with the increasing availability of high frame-rate cameras, video applications, and increasing computational power, direct methods are demonstrating great promise~\cite{lsdslam,kerl13icra,DTAM}. For instance, direct estimation of motion from RGB-D data was shown to be robust, precise and efficient~\cite{kerl13icra,klose2013efficient,steinbrucker2011real}.

To date, however, the use of direct methods in VSLAM has been limited to frame--frame motion estimation (commonly referred to as visual odometry). Approaches that make use of multiple frames are designed for dense depth estimation only and multi-view stereo~\cite{DTAM,mvs}, which assume a correct camera pose and only refine the scene structure. Other algorithms can include measurements from multiple frames, but rely on the presence of structures with strong planarity in the environment~\cite{Irani2000,smr08-itro} (or equivalently assuming a rotation only motion such that the motion of the camera can be represented as a homography~\cite{lovegrove2010real}).

In this work, in contrast to previous research in direct image-based alignment~\cite{DTAM,SteinShashua2000}, we show that provided good initialization, it is possible to jointly refine the structure and motion parameters by minimizing the photometric error and without restricting the camera motion or the scene structure.

The LSD-SLAM algorithm~\cite{lsdslam} is a well-known recently proposed direct algorithm for vision-based motion estimation. In comparison to our work, the fundamental difference is that we refine the parameters of motion and structure jointly in one large optimization problem. In LSD-SLAM, the photometric error is used to estimate the motion, while scene (inverse) depth is estimated using small baseline stereo with fixed camera fixed. The joint optimization of motion and structure proposed herein is important in future work concerning the optimality and convergence properties of photometric structure-from-motion (SFM) and photometric, or direct, VSLAM. Our work can be regarded as an extension of LSD-SLAM where the parameters of motion and structure are refined jointly.

\subsubsection{Dense multi-view stereo (MVS)}
MVS algorithms aim at recovering a dense depth estimate of objects or scenes using many images with known pose~\cite{mvs}. To date, however, research on simultaneous refinement of motion and depth from multiple frames remains sparse. Furukawa and Ponce~\cite{furukawa2008accurate} were among the first to demonstrate that relying on minimizing the reprojection error is not always accurate enough. The work demonstrates that calibration errors could have a large impact on accuracy. Furukawa and Ponce address this problem by refining the correspondences using photometric information and visibility information from an intermediate dense reconstruction in a guided matching step. The process is then interleaved with traditional geometric BA using the improved correspondences to obtain a better reconstruction accuracy. In our work, however, we show that interleaving the minimization of the photometric error with the reprojection error may be unnecessary and solving the problem directly is feasible.

Recently, Delaunoy and Pollefeys~\cite{delaunoy2014photometric} proposed a photometric BA approach for dense MVS. Starting from a precise initial reconstruction and a mesh model of the object, the algorithm is demonstrated to enhance MVS accuracy. The imaging conditions, however, are ideal and brightness constancy is assumed~\cite{delaunoy2014photometric}. In our work, we do not require a very precise initialization and can address challenging illumination conditions. More importantly, the formulation proposed by Delaunoy and Pollefeys requires the availability of an accurate dense mesh, which is not possible to obtain in VSLAM scenarios. Furthermore, initialization requirements appear to be much higher than our approach.

\section{Conclusions}
In this work, we show how to improve on the accuracy of the state-of-art VSLAM methods by minimizing the photometric error across multiple views. In particular, we show that it is possible to improve results obtained by minimizing the reprojection error in a bundle adjustment (BA) framework. We also show, contrary to previous image-based minimization work~\cite{SteinShashua2000,DTAM,steinbrucker2011real,lsdslam,engel2015_stereo_lsdslam}, that the \emph{joint} refinement of motion and structure is possible in unconstrained scenes without the need for alternation or disjoint optimization.

The accuracy of minimizing the reprojection using traditional BA is limited by the precision and accuracy of feature localization and matching. In contrast, our approach --- BA without correspondences --- determines the correspondences implicitly such that the photometric consistency is maximized as a function of the scene structure and camera motion parameters.

Finally, we show that accurate solutions to geometric problems in vision are not restricted to geometric primitives such as corners and edges, or even planes. We look forward to more sophisticated modeling of the geometry and photometry of the scene beyond the intensity patches used in our work.

\hypersetup{urlcolor=black}
{\footnotesize
\bibliographystyle{splncs}
\bibliography{bib}

\begin{thebibliography}{10}

\bibitem{Sun2010}
Sun, D., Roth, S., Black, M.:
\newblock \href{http://dx.doi.org/10.1109/CVPR.2010.5539939}{Secrets of optical
  flow estimation and their principles}.
\newblock In: Computer Vision and Pattern Recognition (CVPR), 2010 IEEE
  Conference on. (2010)  2432--2439

\bibitem{SceneFlow}
Vedula, S., Baker, S., Rander, P., Collins, R., Kanade, T.:
\newblock
  \href{https://www.ri.cmu.edu/publication_view.html?pub_id=4772}{Three-Dimensional
  Scene Flow}.
\newblock IEEE Transactions on Pattern Analysis and Machine Intelligence
  \textbf{27} (2005)  475 -- 480

\bibitem{Seitz2006}
Seitz, S., Curless, B., Diebel, J., Scharstein, D., Szeliski, R.:
\newblock \href{http://dx.doi.org/10.1109/CVPR.2006.19}{A Comparison and
  Evaluation of Multi-View Stereo Reconstruction Algorithms}.
\newblock In: Computer Vision and Pattern Recognition, 2006 IEEE Computer
  Society Conference on. Volume~1. (2006)  519--528

\bibitem{mvs}
Furukawa, Y., Hernández, C.:
\newblock Multi-view stereo: A tutorial.
\newblock Foundations and Trends® in Computer Graphics and Vision \textbf{9}
  (2015)  1--148

\bibitem{lsdslam}
Engel, J., Sch{\"{o}}ps, T., Cremers, D.:
\newblock \href{http://dx.doi.org/10.1007/978-3-319-10605-2_54}{{LSD-SLAM:}
  Large-Scale Direct Monocular {SLAM}}.
\newblock In: {ECCV}. (2014)

\bibitem{kerl13icra}
Kerl, C., Sturm, J., Cremers, D.:
\newblock
  \href{http://vision.in.tum.de/_media/spezial/bib/kerl13icra.pdf}{Robust
  Odometry Estimation for RGB-D Cameras}.
\newblock In: Int'l Conf. on Robotics and Automation (ICRA). (2013)

\bibitem{steinbrucker2011real}
Steinbrucker, F., Sturm, J., Cremers, D.:
\newblock \href{http://dx.doi.org/10.1109/ICCVW.2011.6130321}{Real-time visual
  odometry from dense RGB-D images}.
\newblock In: ICCV Workshops, IEEE Int'l Conf.~on Computer Vision. (2011)

\bibitem{Meilland13}
Meilland, M., Comport, A.:
\newblock \href{http://dx.doi.org/10.1109/IROS.2013.6696881}{On unifying
  key-frame and voxel-based dense visual SLAM at large scales}.
\newblock In: Intelligent Robots and Systems (IROS), 2013 IEEE/RSJ
  International Conference on. (2013)  3677--3683

\bibitem{DTAM}
Newcombe, R., Lovegrove, S., Davison, A.:
\newblock \href{http://dx.doi.org/10.1109/ICCV.2011.6126513}{DTAM: Dense
  tracking and mapping in real-time}.
\newblock In: Computer Vision (ICCV), 2011 IEEE International Conference on.
  (2011)  2320--2327

\bibitem{ba}
Triggs, B., Mclauchlan, P.F., Hartley, R.I., Fitzgibbon, A.W.:
\newblock \href{http://www.metapress.com/link.asp?id=PLVCRQ5BX753A2TN}{Bundle
  Adjustment -- A Modern Synthesis}.
\newblock LNCS (2000)

\bibitem{Hartley2006}
Hartley, R., Zisserman, A.:
\newblock \href{http://www.robots.ox.ac.uk/~vgg/hzbook/}{Multiple View Geometry
  in Computer Vision}. Second edn.
\newblock Cambridge University Press (2004)

\bibitem{torr00}
Torr, P., Zisserman, A.:
\newblock \href{http://dx.doi.org/10.1007/3-540-44480-7_19}{Feature Based
  Methods for Structure and Motion Estimation}.
\newblock In: Vision Algorithms: Theory and Practice.
\newblock Springer Berlin Heidelberg (2000)  278--294

\bibitem{kanazawa2001we}
Kanazawa, Y., Kanatani, K.:
\newblock Do we really have to consider covariance matrices for image features?
\newblock In: Computer Vision, 2001. ICCV 2001. Proceedings. Eighth IEEE
  International Conference on. Volume~2., IEEE (2001)  301--306

\bibitem{brooks2001value}
Brooks, M.J., Chojnacki, W., Gawley, D., Van Den~Hengel, A.:
\newblock What value covariance information in estimating vision parameters?
\newblock In: Computer Vision, 2001. ICCV 2001. Proceedings. Eighth IEEE
  International Conference on. Volume~1., IEEE (2001)  302--308

\bibitem{furukawa2008accurate}
Furukawa, Y., Ponce, J.:
\newblock Accurate camera calibration from multi-view stereo and bundle
  adjustment.
\newblock In: Computer Vision and Pattern Recognition, 2008. CVPR 2008. IEEE
  Conference on, IEEE (2008)  1--8

\bibitem{deriche1990accurate}
Deriche, R., Giraudon, G.:
\newblock Accurate corner detection: An analytical study.
\newblock In: Computer Vision, 1990. Proceedings, Third International
  Conference on, IEEE (1990)  66--70

\bibitem{shimizu2001}
Shimizu, M., Okutomi, M.:
\newblock \href{http://dx.doi.org/10.1109/ICCV.2001.937503}{Precise sub-pixel
  estimation on area-based matching}.
\newblock In: ICCV. Volume~1. (2001)  90--97 vol.1

\bibitem{orbslam2015}
Mur{-}Artal, R., Montiel, J.M.M., Tard{\'{o}}s, J.D.:
\newblock \href{http://arxiv.org/abs/1502.00956}{{ORB-SLAM:} a Versatile and
  Accurate Monocular {SLAM} System}.
\newblock CoRR \textbf{abs/1502.00956} (2015)

\bibitem{Milford2012}
Milford, M., Wyeth, G.:
\newblock \href{http://dx.doi.org/10.1109/ICRA.2012.6224623}{SeqSLAM: Visual
  route-based navigation for sunny summer days and stormy winter nights}.
\newblock In: Robotics and Automation (ICRA), 2012 IEEE International
  Conference on. (2012)  1643--1649

\bibitem{Reid2014}
Reid, I.:
\newblock \href{http://dx.doi.org/10.1109/ICARCV.2014.7064267}{Towards semantic
  visual SLAM}.
\newblock In: Control Automation Robotics Vision (ICARCV), 2014 13th
  International Conference on. (2014)  1--1

\bibitem{Moreno2013}
Salas-Moreno, R., Newcombe, R., Strasdat, H., Kelly, P., Davison, A.:
\newblock \href{http://dx.doi.org/10.1109/CVPR.2013.178}{SLAM++: Simultaneous
  Localisation and Mapping at the Level of Objects}.
\newblock In: Computer Vision and Pattern Recognition (CVPR), 2013 IEEE
  Conference on. (2013)  1352--1359

\bibitem{murray1994mathematical}
Murray, R.M., Li, Z., Sastry, S.S., Sastry, S.S.:
\newblock A mathematical introduction to robotic manipulation.
\newblock CRC press (1994)

\bibitem{ma2004invitation}
\href{http://vision.ucla.edu/MASKS/}{Ma, Yi and Soatto, Stefano and Kosecka,
  Jana and Sastry, S. Shankar}:
\newblock An Invitation to 3-D Vision: From Images to Geometric Models.
\newblock Springer Verlag (2003)

\bibitem{hartley2013rotation}
Hartley, R., Trumpf, J., Dai, Y., Li, H.:
\newblock Rotation averaging.
\newblock International journal of computer vision \textbf{103} (2013)
  267--305

\bibitem{civera2008inverse}
Civera, J., Davison, A.J., Montiel, J.M.:
\newblock Inverse depth parametrization for monocular slam.
\newblock Robotics, IEEE Transactions on \textbf{24} (2008)  932--945

\bibitem{zhao2015parallaxba}
Zhao, L., Huang, S., Sun, Y., Yan, L., Dissanayake, G.:
\newblock Parallaxba: bundle adjustment using parallax angle feature
  parametrization.
\newblock The International Journal of Robotics Research \textbf{34} (2015)
  493--516

\bibitem{lk}
Lucas, B.D., Kanade, T.:
\newblock \href{http://www.ri.cmu.edu/publication_view.html?pub_id=2549}{An
  Iterative Image Registration Technique with an Application to Stereo Vision
  (DARPA)}.
\newblock In: Proc.~of the 1981 DARPA Image Understanding Workshop. (1981)
  121--130

\bibitem{horn1981determining}
Horn, B.K., Schunck, B.G.:
\newblock
  \href{http://people.csail.mit.edu/bkph/papers/Optical_Flow_OPT.pdf}{Determining
  optical flow}.
\newblock Artificial intelligence \textbf{17} (1981)  185--203

\bibitem{baker2004lucas}
Baker, S., Matthews, I.:
\newblock
  \href{http://www.ri.cmu.edu/pub_files/pub3/baker_simon_2004_1/baker_simon_2004_1.pdf}{Lucas-kanade
  20 years on: A unifying framework}.
\newblock International Journal of Computer Vision \textbf{56} (2004)  221--255

\bibitem{engel2015_stereo_lsdslam}
Engel, J., Stueckler, J., Cremers, D.:
\newblock \href{http://vision.in.tum.de/research/vslam/lsdslam}{Large-Scale
  Direct SLAM with Stereo Cameras}.
\newblock In: International Conference on Intelligent Robots and Systems
  (IROS). (2015)

\bibitem{Nocedal2006NO}
Nocedal, J., Wright, S.J.:
\newblock \href{http://www.springer.com/us/book/9780387303031}{Numerical
  Optimization}. 2nd edn.
\newblock Springer, New York (2006)

\bibitem{harris}
Harris, C., Stephens, M.:
\newblock A combined corner and edge detector.
\newblock In: Alvey vision conference. Volume~15., Manchester, UK (1988) ~50

\bibitem{rosten2006machine}
Rosten, E., Drummond, T.:
\newblock Machine learning for high-speed corner detection.
\newblock In: Computer Vision--ECCV 2006.
\newblock Springer (2006)  430--443

\bibitem{klt}
Shi, J., Tomasi, C.:
\newblock \href{http://dx.doi.org/10.1109/CVPR.1994.323794}{Good features to
  track}.
\newblock In: Proc.~of Computer Vision and Pattern Recognition (CVPR). (1994)
  593--600

\bibitem{dellaert2000structure}
Dellaert, F., Seitz, S.M., Thorpe, C.E., Thrun, S.:
\newblock Structure from motion without correspondence.
\newblock In: Computer Vision and Pattern Recognition, 2000. Proceedings. IEEE
  Conference on. Volume~2., IEEE (2000)  557--564

\bibitem{meilland2010}
Meilland, M., Comport, A., Rives, P.:
\newblock \href{http://dx.doi.org/10.1109/IROS.2010.5650380}{A spherical
  robot-centered representation for urban navigation}.
\newblock In: IROS. (2010)

\bibitem{nister04}
Nister, D., Naroditsky, O., Bergen, J.:
\newblock \href{http://dx.doi.org/10.1109/CVPR.2004.1315094}{Visual odometry}.
\newblock In: Computer Vision and Pattern Recognition (CVPR). (2004)

\bibitem{Irani2000}
Irani, M., Anandan, P., Cohen, M.:
\newblock \href{http://dx.doi.org/10.1007/3-540-44480-7_6}{Direct Recovery of
  Planar-Parallax from Multiple Frames}.
\newblock In Triggs, B., Zisserman, A., Szeliski, R., eds.: Vision Algorithms:
  Theory and Practice. Volume 1883 of Lecture Notes in Computer Science.
\newblock Springer Berlin Heidelberg (2000)  85--99

\bibitem{SteinShashua2000}
Stein, G., Shashua, A.:
\newblock \href{http://dx.doi.org/10.1109/34.877522}{Model-based brightness
  constraints: on direct estimation of structure and motion}.
\newblock Pattern Analysis and Machine Intelligence, IEEE Transactions on
  \textbf{22} (2000)  992--1015

\bibitem{agouris1996automated}
Agouris, P., Schenk, T.:
\newblock Automated aerotriangulation using multiple image multipoint matching.
\newblock Photogrammetric Engineering and Remote Sensing \textbf{62} (1996)
  703--710

\bibitem{ceres-solver}
Agarwal, S., Mierle, K., Others:
\newblock \href{http://ceres-solver.org}{Ceres Solver}.
\newblock \url{http://ceres-solver.org} (2016)

\bibitem{levenberg44}
Levenberg, K.:
\newblock \href{http://www.library.cmu.edu/}{A method for the solution of
  certain non-linear problems in least squares}.
\newblock Quart. J. Appl. Maths. \textbf{II} (1944)  164--168

\bibitem{marquardt63}
Marquardt, D.W.:
\newblock
  \href{http://epubs.siam.org/doi/abs/10.1137/0111030?journalCode=smjmap.1}{An
  Algorithm for Least-Squares Estimation of Nonlinear Parameters}.
\newblock Journal of the Society for Industrial and Applied Mathematics
  \textbf{11} (1963)  pp. 431--441

\bibitem{Sunderhauf2006}
Sünderhauf, N., Konolige, K., Lacroix, S., Protzel, P.:
\newblock \href{http://dx.doi.org/10.1007/3-540-30292-1_20}{Visual Odometry
  Using Sparse Bundle Adjustment on an Autonomous Outdoor Vehicle}.
\newblock In Levi, P., Schanz, M., Lafrenz, R., Avrutin, V., eds.: Autonome
  Mobile Systems 2005. Informatik aktuell.
\newblock (2006)  157--163

\bibitem{Geiger2012CVPR}
Geiger, A., Lenz, P., Urtasun, R.:
\newblock
  \href{http://www.webmail.cvlibs.net/publications/Geiger2012CVPR.pdf}{Are we
  ready for Autonomous Driving? The KITTI Vision Benchmark Suite}.
\newblock In: Conference on Computer Vision and Pattern Recognition (CVPR).
  (2012)

\bibitem{blanco2013mlgdataset}
Blanco, J.L., Moreno, F.A., Gonz{\'{a}}lez-Jim{\'{e}}nez, J.:
\newblock The m{\'{a}}laga urban dataset: High-rate stereo and lidars in a
  realistic urban scenario.
\newblock International Journal of Robotics Research \textbf{33} (2014)
  207--214

\bibitem{tardif2010new}
Tardif, J.P., George, M., Laverne, M., Kelly, A., Stentz, A.:
\newblock A new approach to vision-aided inertial navigation.
\newblock In: Intelligent Robots and Systems (IROS), 2010 IEEE/RSJ
  International Conference on, IEEE (2010)  4161--4168

\bibitem{badino2013visual}
Badino, H., Yamamoto, A., Kanade, T.:
\newblock
  \href{https://www.ri.cmu.edu/pub_files/2013/12/badino_cvad13.pdf}{Visual
  odometry by multi-frame feature integration}.
\newblock In: Computer Vision Workshops (ICCVW), 2013 IEEE International
  Conference on. (2013)  222--229

\bibitem{lindeberg1993scale}
Lindeberg, T.:
\newblock \href{http://dx.doi.org/10.1007/978-1-4757-6465-9}{Scale-space theory
  in computer vision}.
\newblock Springer (1994)

\bibitem{bitplanes_vo}
Alismail, H., Browning, B., Lucey, S.:
\newblock Direct visual odometry using bit-planes.
\newblock CoRR \textbf{abs/1604.00990} (2016)

\bibitem{delaunoy2014photometric}
Delaunoy, A., Pollefeys, M.:
\newblock Photometric bundle adjustment for dense multi-view 3d modeling.
\newblock In: Computer Vision and Pattern Recognition (CVPR), 2014 IEEE
  Conference on, IEEE (2014)  1486--1493

\bibitem{unser1999splines}
Unser, M.:
\newblock Splines: A perfect fit for signal and image processing.
\newblock Signal Processing Magazine, IEEE \textbf{16} (1999)  22--38

\bibitem{farid2004differentiation}
Farid, H., Simoncelli, E.P.:
\newblock Differentiation of discrete multidimensional signals.
\newblock Image Processing, IEEE Transactions on \textbf{13} (2004)  496--508

\bibitem{sgm}
Hirschmuller, H.:
\newblock \href{http://dx.doi.org/10.1109/CVPR.2005.56}{Accurate and efficient
  stereo processing by semi-global matching and mutual information}.
\newblock In: Computer Vision and Pattern Recognition. (2005)

\bibitem{bal2010}
Agarwal, S., Snavely, N., Seitz, S., Szeliski, R.:
\newblock \href{http://dx.doi.org/10.1007/978-3-642-15552-9_3}{Bundle
  Adjustment in the Large}.
\newblock In Daniilidis, K., Maragos, P., Paragios, N., eds.: Computer Vision
  – ECCV 2010. Volume 6312 of Lecture Notes in Computer Science.
\newblock (2010)  29--42

\bibitem{konolige2010sparse}
Konolige, K., Garage, W.:
\newblock
  \href{https://www.willowgarage.com/sites/default/files/ssba.pdf}{Sparse
  Sparse Bundle Adjustment.}
\newblock In: BMVC. (2010)  1--11

\bibitem{Jeong2010}
Jeong, Y., Nister, D., Steedly, D., Szeliski, R., Kweon, I.S.:
\newblock Pushing the envelope of modern methods for bundle adjustment.
\newblock IEEE Transactions on Pattern Analysis and Machine Intelligence
  \textbf{34} (2012)  1605--1617

\bibitem{Engels06bundleadjustment}
Engels, C., Stewénius, H., Nister, D.:
\newblock Bundle adjustment rules.
\newblock In: In Photogrammetric Computer Vision. (2006)

\bibitem{wu2011multicore}
Wu, C., Agarwal, S., Curless, B., Seitz, S.M.:
\newblock Multicore bundle adjustment.
\newblock In: Computer Vision and Pattern Recognition (CVPR), 2011 IEEE
  Conference on, IEEE (2011)  3057--3064

\bibitem{Ni2007}
Ni, K., Steedly, D., Dellaert, F.:
\newblock Out-of-core bundle adjustment for large-scale 3d reconstruction.
\newblock In: 2007 IEEE 11th International Conference on Computer Vision.
  (2007)  1--8

\bibitem{Konolige2008}
Konolige, K., Agrawal, M.:
\newblock \href{http://dx.doi.org/10.1109/TRO.2008.2004832}{FrameSLAM: From
  Bundle Adjustment to Real-Time Visual Mapping}.
\newblock Robotics, IEEE Transactions on \textbf{24} (2008)  1066--1077

\bibitem{Kaess2011}
Kaess, M., Ila, V., Roberts, R., Dellaert, F.:
\newblock \href{http://dx.doi.org/10.1007/978-3-642-17452-0_10}{The Bayes Tree:
  An Algorithmic Foundation for Probabilistic Robot Mapping}.
\newblock In Hsu, D., Isler, V., Latombe, J.C., Lin, M., eds.: Algorithmic
  Foundations of Robotics IX. Volume~68 of Springer Tracts in Advanced
  Robotics.
\newblock (2011)  157--173

\bibitem{Kaess08tro}
Kaess, M., Ranganathan, A., Dellaert, F.:
\newblock \href{http://people.csail.mit.edu/kaess/isam/}{{iSAM}: Incremental
  Smoothing and Mapping}.
\newblock IEEE Trans. on Robotics (TRO) \textbf{24} (2008)  1365--1378

\bibitem{Kahl2008}
Kahl, F., Agarwal, S., Chandraker, M.K., Kriegman, D., Belongie, S.:
\newblock Practical global optimization for multiview geometry.
\newblock International Journal of Computer Vision \textbf{79} (2008)  271--284

\bibitem{hartley13}
Hartley, R., Kahl, F., Olsson, C., Seo, Y.:
\newblock Verifying global minima for l2 minimization problems in multiple view
  geometry.
\newblock International Journal of Computer Vision \textbf{101} (2013)

\bibitem{Aftab2015}
Aftab, K., Hartley, R.:
\newblock Lq-bundle adjustment.
\newblock In: Image Processing (ICIP), 2015 IEEE International Conference on.
  (2015)  1275--1279

\bibitem{irani1999direct}
Irani, M., Anandan, P.:
\newblock \href{http://dx.doi.org/10.1007/3-540-44480-7_18}{About Direct
  Methods}.
\newblock In: Vision Algorithms: Theory and Practice. (2000)  267--277

\bibitem{Horn1988}
Horn, B.K.P., Weldon, E.J.:
\newblock {Direct methods for recovering motion} (1988)

\bibitem{Oliensis2000}
Oliensis, J.:
\newblock \href{http://dx.doi.org/10.1109/ICPR.2000.905565}{Direct multi-frame
  structure from motion for hand-held cameras}.
\newblock In: Pattern Recognition, 2000. Proceedings. 15th International
  Conference on. Volume~1. (2000)  889--895 vol.1

\bibitem{Mandelbaum99}
Mandelbaum, R., Salgian, G., Sawhney, H.:
\newblock \href{http://dx.doi.org/10.1109/ICCV.1999.791270}{Correlation-based
  estimation of ego-motion and structure from motion and stereo}.
\newblock In: Computer Vision, 1999. The Proceedings of the Seventh IEEE
  International Conference on. Volume~1. (1999)  544--550 vol.1

\bibitem{klose2013efficient}
Klose, S., Heise, P., Knoll, A.:
\newblock
  \href{http://www6.in.tum.de/Main/Publications/Klose2013a.pdf}{Efficient
  compositional approaches for real-time robust direct visual odometry from
  RGB-D data}.
\newblock In: IEEE/RSJ Int'l Conf.~on Intelligent Robots and Systems. (2013)

\bibitem{smr08-itro}
Silveira, G., Malis, E., Rives, P.:
\newblock \href{http://dx.doi.org/10.1109/TRO.2008.2004829}{An efficient direct
  approach to visual SLAM}.
\newblock IEEE Transactions on Robotics (2008)

\bibitem{lovegrove2010real}
Lovegrove, S., Davison, A.J.:
\newblock Real-time spherical mosaicing using whole image alignment.
\newblock In: European Conference on Computer Vision, Springer (2010)  73--86

\end{thebibliography}
}

\end{document}